\documentclass[journal,comsoc]{IEEEtran}
\usepackage[utf8]{inputenc}
\usepackage{cite}
\usepackage{longtable}
\usepackage{supertabular}
\usepackage{pdfpages}
\usepackage{enumitem}
\usepackage{array}
\usepackage{supertabular}
\usepackage{pdfpages}
\usepackage{algorithm,algorithmic}
\usepackage{tikz}
\usepackage{comment}
\def\checkmark{\tikz\fill[scale=0.4](0,.35) -- (.25,0) -- (1,.7) -- (.25,.15) -- cycle;} 
\title{Federated Learning for 6G: Paradigms, Taxonomy, Recent Advances and Insights}
\author{\IEEEauthorblockN{Maryam Ben Driss$^{\bullet}$,~\IEEEmembership{Student Member,~IEEE,} Essaid Sabir$^{\bullet}$,~\IEEEmembership{Senior Member,~IEEE,} Halima Elbiaze$^{\bullet}$ ,~\IEEEmembership{Senior Member,~IEEE,} Walid Saad$^{*}$,~\IEEEmembership{Fellow Member,~IEEE,}}\\
\IEEEauthorblockA{$^{\circ}$Department of Computer Science, University of Quebec at Montreal (UQAM), Montreal, H2L 2C4, Canada}\\
\IEEEauthorblockA{$^{*}$Wireless\@VT, Bradley Department of Electrical and Computer Engineering, Virginia Tech, Arlington, VA, USA}
}
\date{September 20th, 2023}
\begin{document}
\maketitle
\begin{abstract}
Artificial Intelligence (AI) is expected to play an instrumental role in the next generation of wireless systems, such as sixth-generation (6G) mobile network. However, massive data, energy consumption, training complexity, and sensitive data protection in wireless systems are all crucial challenges that must be addressed for training AI models and gathering intelligence and knowledge from distributed devices. Federated Learning (FL) is a recent framework that has emerged as a promising approach for multiple learning agents to build an accurate and robust machine learning models without sharing raw data. By allowing mobile handsets and devices to collaboratively learn a global model without explicit sharing of training data, FL exhibits high privacy and efficient spectrum utilization. While there are a lot of survey papers exploring FL paradigms and usability in 6G privacy, none of them has clearly addressed how FL can be used to improve the protocol stack and wireless operations. The main goal of this survey is to provide a comprehensive overview on FL usability to enhance mobile services and enable smart ecosystems to support novel use-cases. This paper examines the added-value of implementing FL throughout all levels of the protocol stack. Furthermore, it presents important FL applications, addresses hot topics, provides valuable insights and explicits guidance for future research and developments. Our concluding remarks aim to leverage the synergy between FL and future 6G, while highlighting FL's potential to revolutionize wireless industry and sustain the development of cutting-edge mobile services.
\end{abstract} 

\section{Introduction}
\subsection{Artificial Intelligence for Wireless}
Artificial Intelligence (AI) is expected to play a prominent role in current and future wireless systems, such as fifth-generation (5G) and sixth-generation (6G) \cite{saad2019vision,jiang2021road}. Indeed, 6G systems are envisioned to be AI-native systems in which some form of AI will be implemented for automation across the protocol stack \cite{chaccour2022less,dahlman2019artificial,letaief2019roadmap}. For instance, AI is expected to provide new frameworks for improving wireless performance metrics, including capacity, latency, efficiency, power, spectrum frequency, flexibility, compatibility, and quality of experience \cite{chen2019artificial}. AI techniques will influence next-generation network systems, due to two key reasons:
\begin{itemize}
    \item \emph{Need for more autonomy:} Machine Learning (ML) techniques are necessary to endow wireless systems with autonomy and self-dependence by analyzing real-time data, such as traffic patterns, user behavior, and environmental factors, to make intelligent decisions for optimizing network resources, routing, and load balancing. This level of dynamic network management allows wireless systems to adapt to changing conditions without manual intervention.
    \item \emph{Need for more communication:} AI systems, with their ability to process vast amounts of data and learn from interactions, significantly contribute to making communication more accessible, efficient, and personalized to meet the diverse and growing demands of modern users and applications.
\end{itemize}
Numerous recent research activities have been focused on surveying the core wireless networking issues \cite{8755300} that are addressed using various ML techniques such as spectrum management and sharing \cite{rastegardoost2018machine,liu2017dynamic,jiang2011efficient,7343323,7462490,7036917}, resource allocation \cite{gai2018optimal,8329631,8403505,ray2018reverse,8501524,challita2017deep}, fading channels prediction \cite{jiang2018multi,jiang2020deep,jiang2020deep4,jiang2020recurrent,konstantinov2019fading}, and traffic prediction \cite{article,5507338,suthaharan2014big,kong2019big}. 
However, applying traditional ML schemes in large-scale systems is still challenging \cite{verbraeken2020survey}\cite{kato2020ten}, for the following reasons (non-exhaustive list):
\begin{itemize}
    \item \emph{Scalability:} Large-scale systems often handle massive amounts of data and require complex computations. In a centralized approach, all data is collected and processed on a single server, resulting in performance bottlenecks and increased processing times.
    \item \emph{Communication overhead:} In a centralized scheme, data from various sources must be transmitted to the central server for processing. This leads to significant communication overhead, especially when dealing with geographically distributed data sources.
    \item \emph{Single point of failure:} Centralized systems have a single point of failure. Any failure of the server will impact the entire ML process. This vulnerability is critical for real-time or mission-critical applications that require continuous and reliable operation.
    \item \emph{Privacy and security concerns:} Centralized ML involves collecting and storing data from multiple sources in a central location. This raises concerns about privacy and security, as sensitive data are at risk without proper security measures in place.
    \item \emph{Latency and real-time requirements:} For real-time or low-latency applications, a centralized approach is not feasible due to the long delay of processing and transmission to the central server.
    \item \emph{Cost and infrastructure:} Building and maintaining a centralized ML infrastructure for large-scale systems is costly. It requires substantial computing power, storage, and network resources to handle the volume of data and computation involved.
\end{itemize}
To address these limitations, the concept of federated learning (FL) has been recently proposed, as an effective approach to train an ML model in a distributed way \cite{shome2022federated}. FL allows multiple devices or nodes to collaboratively train a global model without sharing local data. In traditional ML, data is collected from various sources, sent to a centralized server, and then used to train a global model. However, the training process of FL occurs directly on individual devices or nodes without sharing raw data. FL framework offers a promising approach for future wireless systems by combining the benefits of distributed processing, privacy preservation, scalability, and adaptability. It empowers the evolution of wireless communication and enables the development of innovative and data-driven applications more efficiently and securely.
Still, there are several limitations to implementing FL due to the distributed and dynamic nature of wireless networks. Some of the key challenges include:
\begin{itemize}
    \item \emph{Limited bandwidth:} Transmitting model updates between the central server and edge devices is resource-intensive and leads to increased communication overhead.
    \item \emph{Unreliable connectivity:} Managing the participation of devices with varying connectivity and capabilities requires robust mechanisms for synchronization and communication.
    \item \emph{Computing constraints:} Participating clients are usually equipped with limited computational capabilities, which may affect the convergence speed and whole model performance.
    \item \emph{Imbalance and non-IID data:} In wireless networks, the data collected by edge devices is generally imbalanced or not independent and identically distributed (non-IID), which presents challenges for model convergence and generalization.
\end{itemize}
Addressing these challenges requires a careful combination of algorithmic advancements, optimization techniques, and architectural design to make FL feasible and effective in wireless environments. As the field of FL evolves, researchers continue to work on overcoming these challenges to enable practical and secure deployments in future wireless networks.
\subsection{Relevant Surveys and our Contribution}
\begin{figure}
\centering
\includegraphics[scale=0.4467]{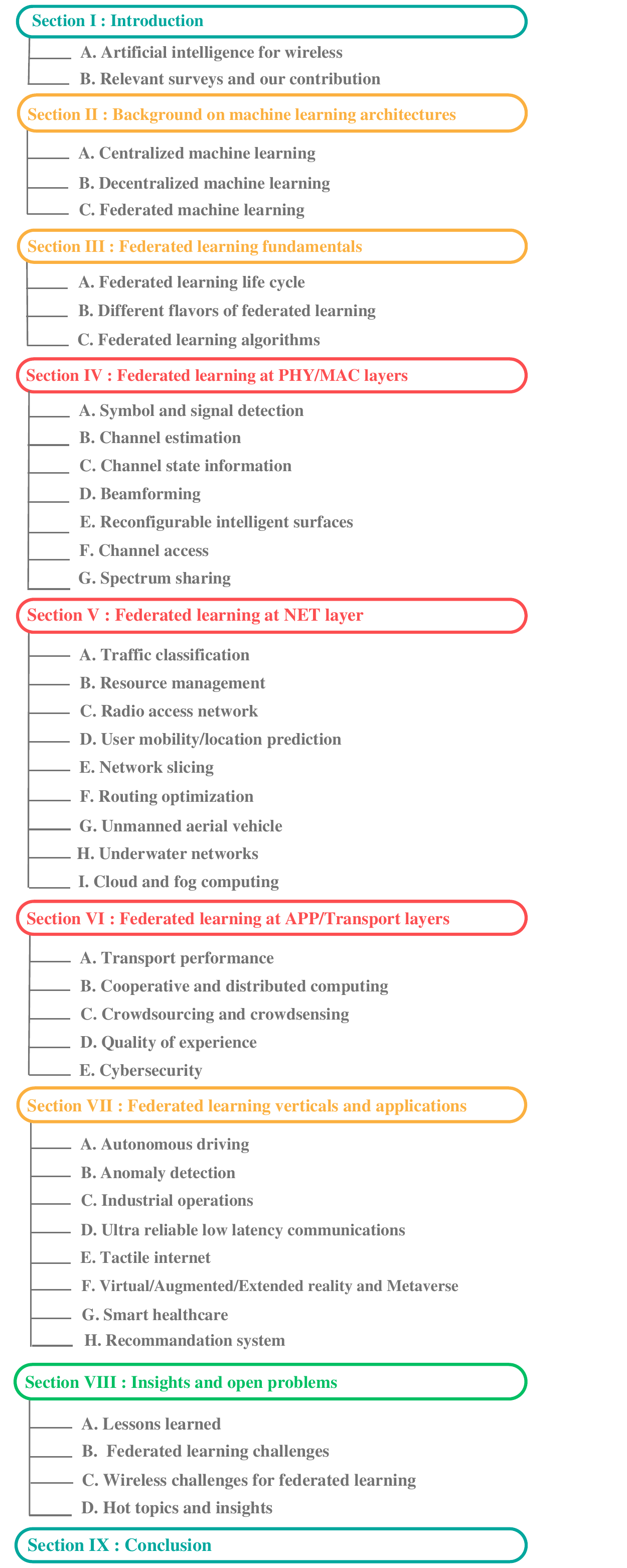}
\caption{The survey paper organization.}
\label{plan}
\end{figure}
Although there are many surveys on FL and wireless networks \cite{lim2020federated,niknam2020federated,hu2020distributed,qin2020federated,mollel2021survey,yang2021federated,elbir2021federated,wahab2021federated,nguyen2021federated,khan2021federated,gadekallu2021federated,al2021edge,kontar2021internet,liu2021distributed,liu2021distributed,shome2021federated,rodriguez2022survey,banabilah2022federated,agrawal2022federated,ghimire2022recent,pandya2023federated,wen2023survey,al2023edge,zhao2023towards} as summarized in Table \ref{table:0}, they do not address their applicability across the protocol stack or provide a detailed examination of FL applications in each layer of the Open Systems Interconnection (OSI) model. This survey aims to fill this gap by providing an in-depth overview of how FL is implemented throughout the entire protocol stack. We also highlight associated advantages and discuss induced challenges. State-of-the-art contributions that utilize federated ML schemes to address problems in wireless communications are collected and discussed. Our main contributions include:
\begin{enumerate}
    \item Drawing a comprehensive background on recent FL paradigms, taxonomy and main techniques;
    \item Comparing centralized ML, distributed ML, and FL in the context of massive and sensitive environments while highlighting the characteristics and the issues associated with each approach; 
    \item Exploring the latest FL schemes aiming to solve the major issues at the physical (PHY), medium access control (MAC), network (NET), transport, and application (APP) layers while achieving high accuracy, enhanced communication efficiency and low energy consumption. We also draw a detailed analysis of the benefits and drawbacks of each application/solution;
    \item Discussing relevant insights and open issues related to the application of FL in 6G and future mobile networks.
\end{enumerate}
For a pleasant reading experience, our survey is organized as depicted in Fig. \ref{plan}. Section II presents ML architectures and their challenges in the context of wireless systems. In Section III, we present a comprehensive overview of FL fundamentals, features, taxonomy, frameworks, benefits, and issues. Next, we explore FL for wireless operations, while focusing on PHY/MAC, NET and Transport/APP layers, in Section IV, Section V and Section VI, respectively. Next, we present recent FL verticals and applications in Section VII. Then, Section VIII summarizes the lessons learned and provide a detailed discussion of FL-wireless open problems that need to be addressed for successful integration in 6G and future systems. Finally, we highlight some promising directions for future research towards efficient AI-native networking. 
{\renewcommand{\arraystretch}{1.5}%
\begin{table*}[ht]
\centering
\begin{tabular}{||p{0.6cm}|l|l|c|c|c|c|c|c||}
 \hline
 \textbf{Ref} & \textbf{Year} & \textbf{Target application} & \textbf{Basics} & \textbf{PHY Layer} & \textbf{NET layer} & \textbf{Transport Layer} & \textbf{APP Layer} & \textbf{Challenges} \\ [0.5ex] 
 \hline
 
 \cite{lim2020federated}
 & 2020 & FL in mobile edge networks & \checkmark & \checkmark & & \checkmark & & \checkmark\\
 
 \cite{niknam2020federated}
 & 2020 & FL for wireless communications & \checkmark & \checkmark & & & & \checkmark\\

 \cite{hu2020distributed} 
 & 2020 & Distributed ML for communication networks & \checkmark & \checkmark & & & \checkmark & \checkmark\\
 
 \cite{qin2020federated} 
 & 2020 & FL and wireless communications & \checkmark & & & & \checkmark & \checkmark\\
 
 \cite{mollel2021survey}
 & 2021 & ML and FL for handover management in 5G & \checkmark & & \checkmark & & & \checkmark\\
 
 \cite{yang2021federated} 
  & 2021 & Federated machine learning for 6G & \checkmark & \checkmark & & & \checkmark & \\
 
 \cite{elbir2021federated} 
 & 2021 & FL for physical layer & \checkmark & \checkmark & & & & \\
 
 \cite{wahab2021federated}
 & 2021 & FL in networking systems & \checkmark & & \checkmark & & & \checkmark\\
 
 \cite{nguyen2021federated}
  & 2021 & Federated learning for IoT & \checkmark & & \checkmark & & \checkmark &\checkmark\\
 
 \cite{khan2021federated}
 & 2021 & Recent advances of FL for IoT networks & \checkmark & \checkmark & & & \checkmark & \checkmark\\
 
 \cite{gadekallu2021federated}
 & 2021 & Federated ML for big data & \checkmark & & & & \checkmark & \checkmark\\
 
 \cite{al2021edge}
 & 2021 & Intelligence for 6G using FL & \checkmark & \checkmark & & & \checkmark & \checkmark\\
 
 \cite{kontar2021internet}
 & 2021 & The internet of federated things & \checkmark & & & & \checkmark & \checkmark\\
 
 \cite{liu2021distributed}
 & 2022 & From distributed ML to FL & \checkmark & & & & & \\
 
 \cite{shome2021federated}
 & 2022 & FL and next-generation communications & \checkmark & \checkmark & \checkmark & & & \checkmark\\
 
 \cite{rodriguez2022survey}
 & 2022 & FL attacks and threats & \checkmark & & & & & \checkmark \\

 \cite{banabilah2022federated}
 & 2022 & Federated learning fundamentals & \checkmark & & & & & \checkmark \\

 \cite{agrawal2022federated}
  & 2022 & FL for intrusion detection & \checkmark & & & & \checkmark & \checkmark \\

 \cite{ghimire2022recent}
 & 2022 & FL for IoT and cybersecurity & \checkmark & & & & \checkmark & \checkmark \\

 \cite{pandya2023federated}
  & 2022 & The application of FL in smart cities & \checkmark & & & &  & \checkmark \\

  \cite{wen2023survey}
  & 2023 & FL challenges and applications & \checkmark & & & &  & \checkmark \\
  
  \cite{al2023edge}
  & 2023 & Trends and challenges of FL for 6G & \checkmark & & & &  \checkmark & \checkmark \\

  \cite{zhao2023towards}
  & 2023 & Communication efficiency in FL & \checkmark & \checkmark & & &  & \checkmark \\
  
 Our work
 & 2023 & Multilayered survey on FL for B5G and 6G & \checkmark & \checkmark & \checkmark & \checkmark & \checkmark & \checkmark\\
 \hline
\end{tabular}
\vspace*{0.1cm}
\caption{Related existing survey papers on FL for network communications.}
\label{table:0}
\end{table*}}

\section{Machine Learning Architectures}
Here, we provide an overview of ML architectures including traditional centralized learning, distributed learning and FL.
\subsection{Centralized Machine Learning}
\begin{figure}[t]
\centering
\includegraphics[width=\columnwidth]{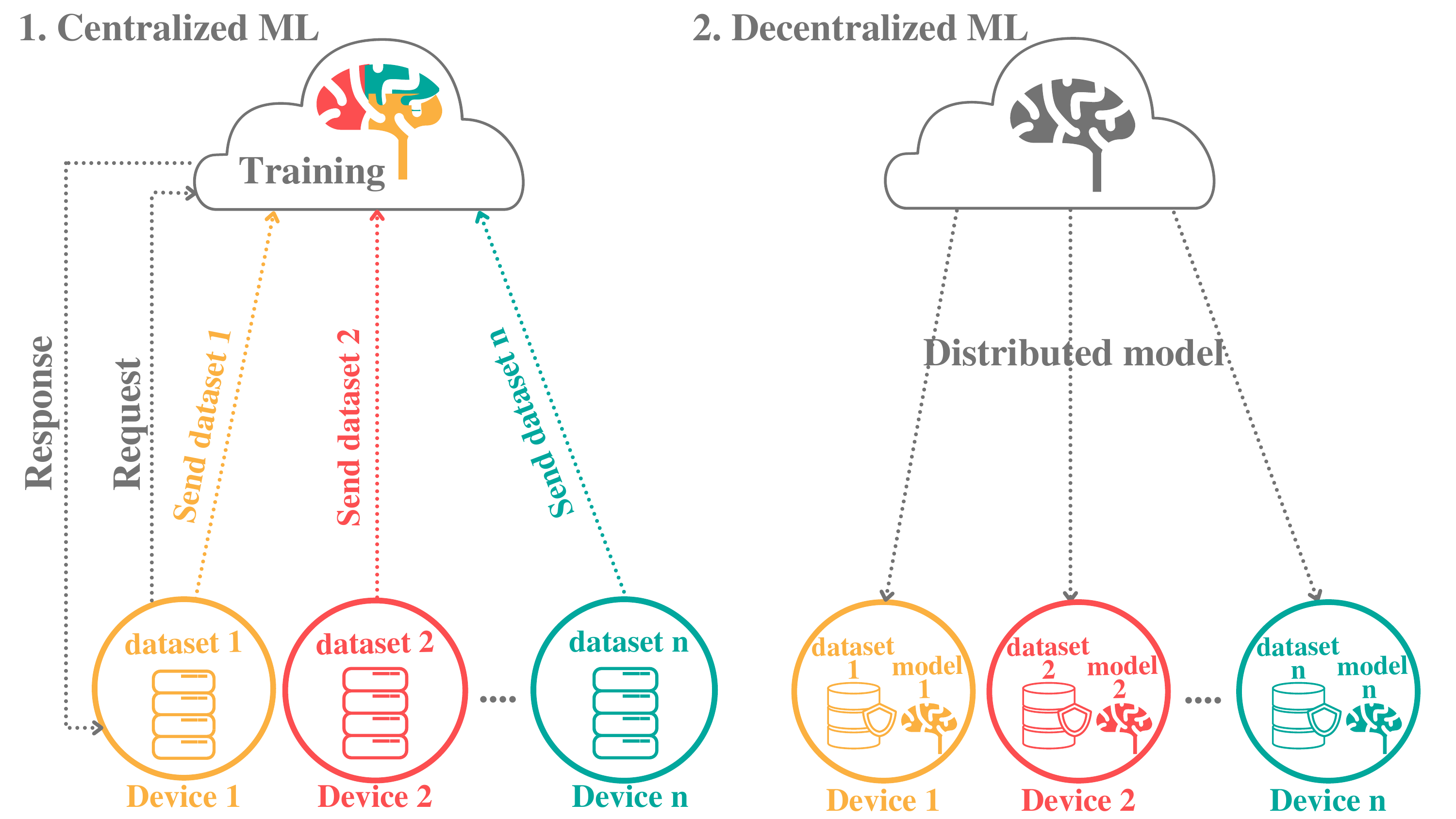}
\caption{Centralized ML has to store data in one data center. Decentralized ML distributes the model across connected devices.}
\label{architecture1}
\end{figure}

ML is the science of enabling computers to learn without being explicitly programmed. This branch of AI allows systems to identify patterns in data, make decisions, and predict future outcomes \cite{boutaba2018comprehensive,zhang2020machine}. Standard ML techniques require data to be stored on a single machine for processing and model training (Fig. \ref{architecture1}.1). Today, Internet of Things (IoT) applications use traditional ML \cite{jagannath2019machine} by uploading data from IoT sensors to the cloud. The cloud server trains a global model using data from multiple devices, allowing immediate interaction with other participants \cite{gunduz2019machine}. However, centralized ML suffers from numerous drawbacks, especially in wireless environments \cite{ma2022privacy}. Hereafter, we provide a non-exhaustive list of these limitations:
\begin{itemize}
    \item \emph{Privacy issues:} Centralized learning involves collecting and storing sensitive data from various sources, raising significant privacy concerns as a potential target for unauthorized access \cite{8940936}. 
    \item \emph{Latency issues:} In centralized learning, data from individual devices is transmitted to a central server. This data transfer introduces latency, which is highly undesirable in applications like real-time communications, autonomous vehicles, or industrial automation, where low latency is critical.
    \item \emph{Costs:} Maintaining and operating centralized servers with the computational power and storage capacity required for large-scale ML is expensive for network operators \cite{sze2017hardware}. 
\end{itemize}
It is worth noting that decentralized approaches, such as edge ML, have been designed to address some of the challenges of centralized ML techniques. 
\subsection{Decentralized Machine Learning}
The concept of decentralized ML was proposed (See \cite{y2018decentralized} and \cite{murshed2019machine}) as an effective solution to address the key challenges listed in Section II.A. As shown in Fig. \ref{architecture1}.2, the decentralized ML approach distributes the computational and data processing tasks across multiple decentralized devices rather than relying on a central server \cite{9220780} which offers several benefits, such as: 
\begin{itemize}
    \item Simplifying learning since each device needs only its dataset while sharing minimal information with external participants.
    \item Adapting to changes over time and learning without being limited by an internet connection or reliance on a central device.
    \item Allowing sensitive data to remain on local devices or edge servers without transmitting to a central location. This enhances data privacy and security.
\end{itemize}
Overall, edge learning empowers wireless communication systems to efficiently handle data, improve response times, enhance privacy, and create more resilient and scalable applications, making it a valuable technology for the growing IoT and 5G ecosystems. Nevertheless, various challenges limit the generated local models to benefit from peer data and learn from external experiences, such as: 
\begin{itemize}
    \item \emph{Data heterogeneity:} In large-scale environments, data collected from different sources has diverse distributions, leading to challenges in combining and aggregating the models effectively.
    \item \emph{Communication overhead:} The exchange of model updates among distributed nodes introduces communication overhead, particularly in bandwidth-constrained networks.
    \item \emph{Synchronization:} Coordinating the timing of model updates across distributed nodes is complex, leading to synchronization issues and potential inconsistencies.
\end{itemize}
\begin{figure}[t]
\centering
\includegraphics[width=\columnwidth]{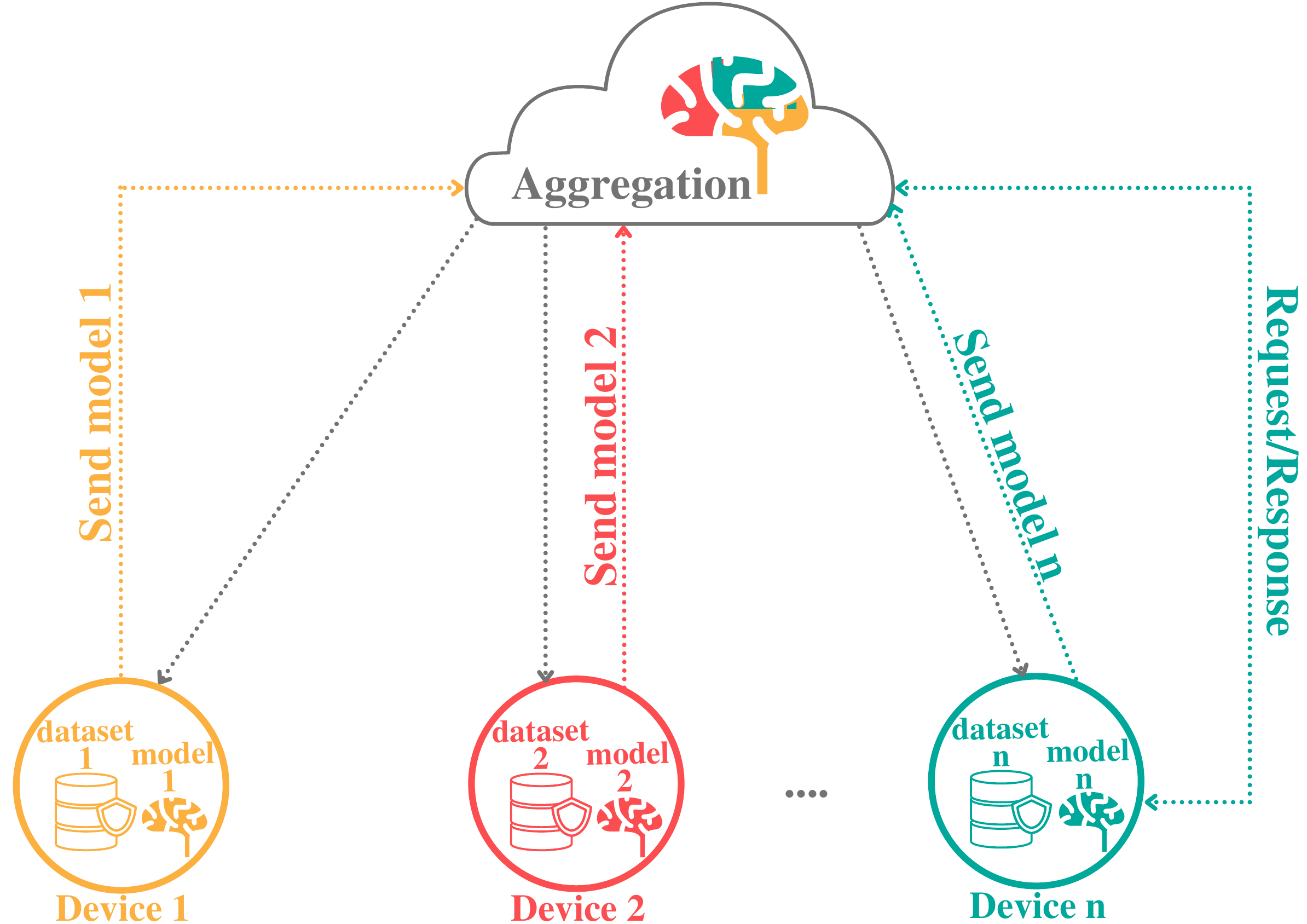}
\caption{FL: Collaborative ML without centralized training data.}
\label{architecture2}
\end{figure}
FL offers a privacy-preserving, scalable, and efficient approach that addresses many of the challenges associated with traditional centralized and decentralized ML, making it a valuable solution in the era of data privacy and decentralization. It is particularly well-suited for applications involving sensitive data, edge computing, and distributed environments.
\subsection{Federated Machine Learning}
\begin{figure}[t]
\centering
\includegraphics[width=6cm]{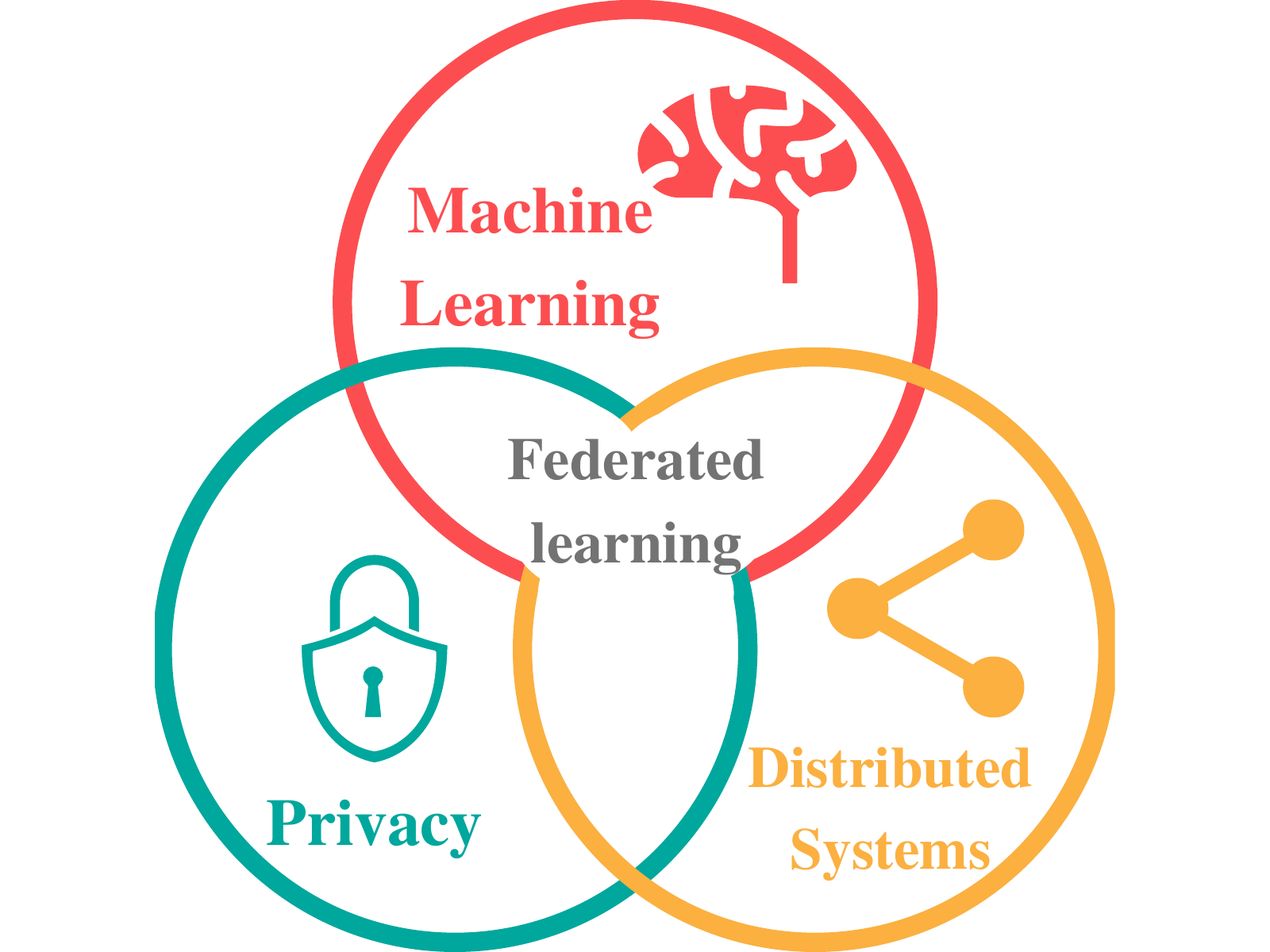}
\caption{FL encompasses three synergistic research areas: ML, privacy, and distributed systems.}
\label{schema}
\end{figure}

FL was recently proposed by Google \cite{konevcny2016federated} as a promising approach for performing distributed ML tasks without relying on a centralized data center. The authors in \cite{DBLP} defined FL as an ML setting wherein multiple entities (clients) work together to solve an ML problem under the coordination of a central server or service provider. In this approach, personal data remains stored locally and is not exchanged or transferred between clients (e.g., see Fig.\ref{architecture2}). Instead, the model updates designed for aggregation are used to achieve the learning objective, enabling secure training while leveraging the collective knowledge of the distributed entities. FL is based on the idea of training a model locally at the data source. The devices communicate their models by combining partial training results into a new supermodel, which is then sent back to all devices \cite{zhang2021survey}. This mechanism is strategically designed to protect privacy and enhance data security \cite{mothukuri2021survey}.
Fig \ref{schema} illustrates that FL depends on the collaboration of ML, privacy, and distributed systems expertise which is fundamental to addressing the challenges posed by distributed data and privacy concerns while harnessing the collective intelligence of decentralized entities for improved model performance and scalability.

\section{Federated Learning Fundamentals}
Before discussing FL applications in wireless systems, it is necessary to provide a brief overview of FL concepts, taxonomy, and techniques. The main concepts of FL are summarized in Table \ref{table:2}.
{\renewcommand{\arraystretch}{1.9}%
\begin{table}[t]
\centering
\begin{tabular}{||m{1.8cm} | p{6cm}||}
    \hline
    \textbf{Goal} & Distributed model training without centralized data collection and with privacy guarantees \tabularnewline
    \hline
    \textbf{Setting} & $K$ devices, out of which $C$ are selected ($C \leq $K) \tabularnewline
    \hline
    \textbf{Parameters} & 
    \begin{itemize}[left=0pt,topsep=0pt]
        \item Batch size
        \item Number of clients
        \item Local iterations
    \end{itemize} \nointerlineskip
    \tabularnewline
    \hline
    \textbf{Orchestration} &
    A central orchestration server or service organizes the training but never sees raw data.
    \tabularnewline
    \hline
    \textbf{Advantages} & 
    \begin{itemize}[left=0pt,topsep=0pt]
        \item Respecting user's privacy
        \item Minimum hardware required
        \item Saving user’s resources
        \item Working offline 
        \item Computing in real-time 
        \item Decreasing training complexity
    \end{itemize} \nointerlineskip\\ 
    \hline
\end{tabular}
\vspace*{0.1cm}
\caption{FL concepts.}
\label{table:2}
\end{table}
}
\subsection{Federated Learning Life Cycle}
\begin{figure}[t]
\centering
\includegraphics[width=6cm]{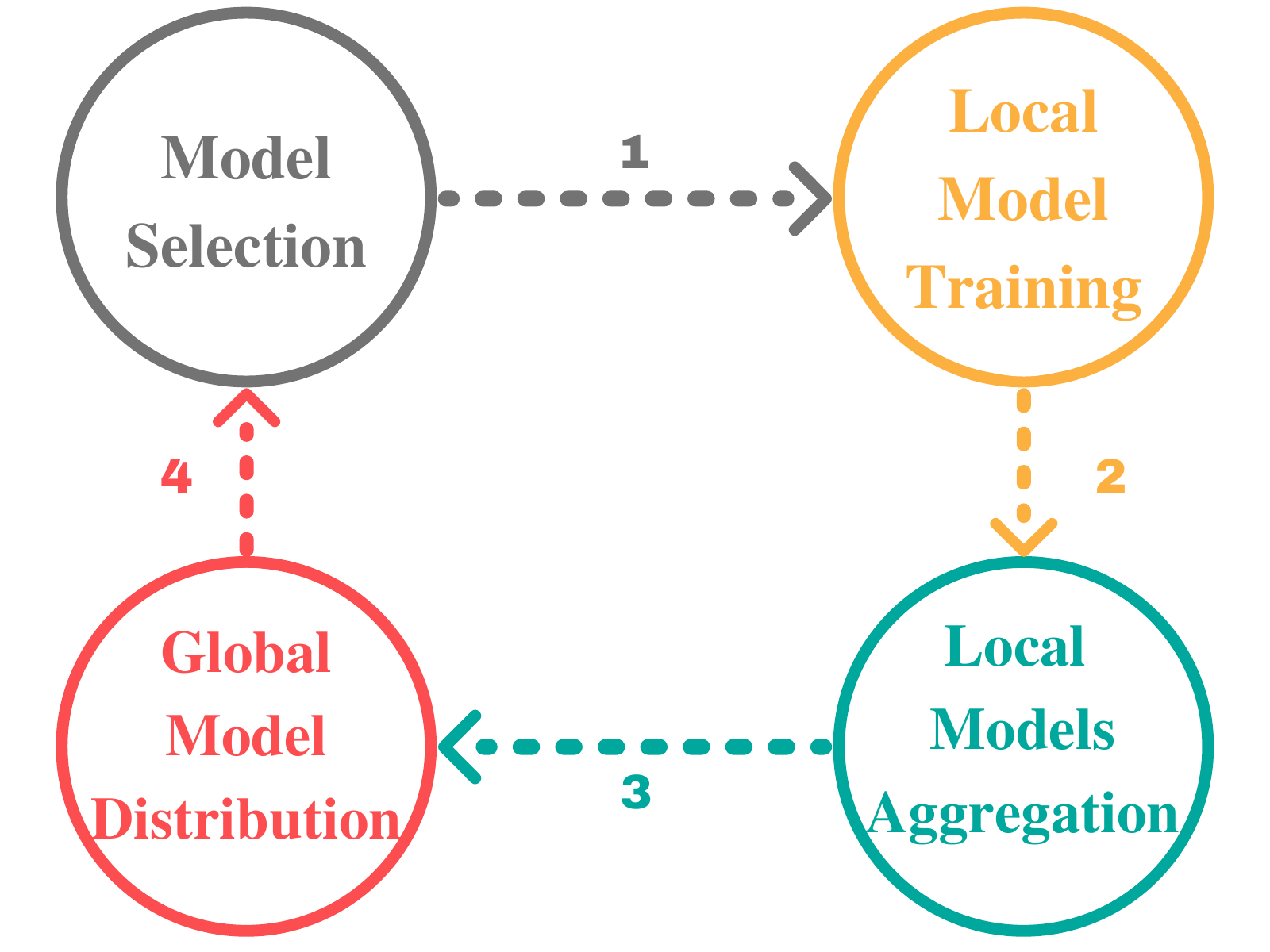}
\caption{FL Life Cycle.}
\label{architecture3}
\end{figure}

The FL process is typically driven by an engineer developing a model for a specific application \cite{DBLP}. In Fig. \ref{architecture3}, the FL workflow generally comprises four steps that occur sequentially.
\begin{itemize}
\item \emph{Model selection:} The global model and parameters are initiated on a central server and shared with all participants.
\item \emph{Local model training:} Clients train the model locally using their data without sharing it with the central server or other clients.
\item \emph{Aggregation of local models:} The update is sent to the central server to aggregate received parameters and create a new global model.
\item \emph{Distribution of global model:} The distribution of the global model among participants for the next iteration.
\end{itemize}
\subsection{Different Flavors of FL}
FL frameworks are classified into three types: horizontal, vertical, and federated transfer learning, based on the distribution of data samples and features \cite{yang2019federated,li2020review,hu2020distributed}.
\subsubsection{Horizontal Federated Learning}
Horizontal FL (HFL) or sample-based FL is applied to scenarios in which datasets share the same feature space but differ in the sample \cite{yang2019federated}. In other words, the participating client devices share the same features but target different populations. considering two banks operating in the same country, the clients of those banks are non-overlapping, their data are likely to have a similar feature space since they adopt similar businesses and operate in the same country. Partitioning by samples is usually relevant when a single company cannot centralize its data due to legal constraints or when organizations with similar aims want to improve their models collaboratively \cite{kairouz2019advances}. Wake-word recognition \cite{leroy2019federated}, such as "Hey Siri" and "OK Google", is a typical application of horizontal partition since each user says the same sentence in a different accent.
\subsubsection{Vertical Federated Learning}
Vertical FL (VFL) is specifically applicable in scenarios where two datasets share the same sample space but differ in their feature space \cite{yang2019federated}. To illustrate this FL category, consider two companies located in the same town, one being a bank, and the other an e-commerce company. While their user base is overlapping, their data collections diverge. The bank tracks user sales, spending activity, and credit scores, while the e-commerce store captures user search and buying history. As the objective is to train a model predicting product purchases based on user features, VFL emerges as the ideal solution to gather additional features and construct a comprehensive model using data from both companies.

\subsubsection{Federated Transfer Learning}
Federated transfer learning (FTL) is suitable where two datasets vary in terms of samples and features space \cite{liu2020secure}. Considering two companies from two countries whose users and feature spaces have a small intersection. In this scenario, FTL learns from the common instances between the spaces using the limited standard sample sets. The knowledge gained is later applied to make predictions for samples with only one-sided features. In wireless systems, FTL enables devices to remember what they have learned and what other devices have learned \cite{liu2019lifelong} and this is applicable to cases in which multiple wireless devices make decisions in different environments.

\subsubsection{Other Categories}
There are other FL architectures, beyond HFL, VFL, and FTL. For instance, multi-participant multi-class VFL (MMVFL) was proposed in \cite{feng2020multi} to manage many users. It allows privacy-preserving label distribution from its owner to other participants. FEDF is another framework introduced in \cite{cao2020federated} to train ML models on multiple geographically distributed training datasets belonging to several owners. Another FL framework, called PerFit, was developed in \cite{wu2020personalized} to enhance the performance of devices in IoT applications by learning a shared model through aggregating local updates from distributed devices and utilizing the benefits of edge computing. This approach enables personalization and improved performance for IoT devices. In general, there are several architectures for FL depending on the distribution characteristics of the data (e.g., see Table \ref{table:3}).
{\renewcommand{\arraystretch}{1.5}%
\begin{table*}[ht]
\centering
\begin{tabular}{||l |l |l |l |l ||}
 \hline
 \textbf{Ref} & \textbf{Architecture} & \textbf{Focuses} &  \textbf{Benefits} & \textbf{Drawbacks} \\ [0.5ex] 
 \hline
 \cite{yang2019federated} & HFL
  & Security & Independence & Need to provide more security \\
 \cite{yang2019federated} & VFL
  & Privacy & Encryption & Handling only two participants\\
 \cite{liu2020secure} & FTL
  & Avoiding accuracy loss & High precision & Expensive computation\\
 \cite{feng2020multi} & MMVFL 
  & Data leakage &  Multiple participants &  Need to handle stragglers effectively\\
 \cite{cao2020federated} & FEDF 
  & Parallel training & Accelerating training & Enormous data exchange \\ 
 \cite{wu2020personalized} & PerFit 
  & IOT applicability & Cloud based & Data augmentation step is required \\ 
 \cite{huang2021starfl} & StarFL 
  & Urban computing & Security &  System efficiency \\
 \hline
\end{tabular}
\vspace*{0.1cm}
\caption{FL architectures: focuses, benefits and drawbacks.}
\label{table:3}
\end{table*}}
Furthermore, the hybrid federated centralized learning (HFCL) approach proposed in \cite{elbir2020hybrid} contributes to the learning process with their datasets. In cases where certain edge devices lack the computational capacity to compute gradients during model training, HFCL proves valuable because it exclusively enables active devices with sufficient computing capabilities to conduct gradient computations on their respective local datasets and engage in collaborative training of the learning model. In contrast, the remaining passive devices send their local datasets to the parameter server.
In \cite{MM}, the authors proposed a novel concept of collaborative FL, whereby devices leverage FL techniques without relying on a centralized parameter server. The work in \cite{MM} showcases the architecture and challenges of this CFL paradigm.

\subsection{Federated Learning Algorithms}
In FL, an ML model is learned iteratively by repeating the following steps: 1) The server chooses a set of users to compute an updated model; 2) Each client computes a locally trained model to update its local data; 3) Updates are transferred to the server; and 4) The server aggregates these local updates to construct a global model. The computation happens on the user's devices and the central aggregator, while contact occurs between them \cite{aledhari2020federated}. The goal of computation is the model preparation, and the communication usually aims to share model parameters \cite{li2019survey}. In this section, we present important FL algorithms (Table \ref{table:3_1}).

\subsubsection*{\textbf{FedAvg}} 
Federated averaging (FedAvg) is a popular algorithm that uses an iterative averaging process to train the global model \cite{mcmahan2017communication}. It divides the data into multiple non-overlapping subsets, and each client trains a local model on its subset. The local models are then sent to the server, and their parameters are averaged to create a new global model. This process is repeated for a fixed number of rounds until the global model converges. The steps of FedAvg are almost identical to traditional ML and deep learning (DL) techniques with a parameter-server and workers \cite{mcmahan2017communication}.

\subsubsection*{\textbf{FedSGD}}
The federated stochastic gradient descent is similar to FedAvg, devices use stochastic gradient descent to update models locally, and each device adjusts its learning rate, enabling better convergence for devices with different data distributions.

\subsubsection*{\textbf{FedProx}}
In heterogeneous networks, federated optimization incorporates a proximal term to link the local and global models, to minimize excessive drift during the optimization process. FedProx modifies the FedAvg algorithm by enabling partial work to be conducted across devices based on the underlying system restrictions and utilizing the proximal term to securely include the partial work \cite{li2018federated}.

\subsubsection*{\textbf{FedATT}}
Attentive federated examines the relative significance of local models and aggregates them using a layer-wise soft attention mechanism between local and global models. This attentive aggregation method minimizes the weighted distance between the server and client models on non-IID datasets \cite{ji2019learning}.

\subsubsection*{\textbf{SimFL}}
Similarity-based FL (SimFL) represents an FL framework where no trusted server is needed \cite{li2020practical}. It includes four main steps:
\begin{enumerate}
    \item The parties first update the gradients of their local data. 
    \item The gradients are sent to a selected party. 
    \item The selected party uses its local data and gradients to update the model. 
    \item The model is sent to all the other parties.
\end{enumerate}
To promote fairness and effectively leverage data from multiple parties, each participating party is chosen to update the model for a similar number of rounds and outputs the final model.

Various FL frameworks are designed to facilitate the implementation and management of the FL process. These frameworks provide tools, libraries, and APIs that enable developers to create and deploy FL algorithms:

\subsubsection*{\textbf{TFF}} The tensorflow federated is an open-source framework for ML and other computations on decentralized data. It has been developed to facilitate open research and experimentation with FL.

\subsubsection*{\textbf{PySyft}} A privacy-preserving FL framework built over PyTorch was introduced in \cite{ryffel2018generic}. The design relies on chains of tensors exchanged between local and remote workers.

\subsubsection*{\textbf{LEAF}} LEAF was proposed in \cite{DBLP:journals/corr/abs-1812-01097}; it is a modular framework for learning in federated settings or ecosystems marked by massively distributed networks of devices.

\subsubsection*{\textbf{Paddle FL}} An open-source framework designed mainly for industrial applications \cite{ma2019paddlepaddle} which effectively replicates several FL algorithms for large-scale distributed clusters.

\subsubsection*{\textbf{FATE}} The federated AI technology enabler is an open-source platform that aims to create a safe computing environment for the federated AI ecosystem \cite{FATE}.
\subsubsection*{\textbf{IBM FL}} A python-based FL framework proposed by IBM for enterprise environments, which provides an essential fabric for adding advanced features \cite{ludwig2020ibm}.

\subsubsection*{\textbf{FedML}} An open research library proposed in \cite{he2020fedml} to support the development of FL algorithms and fair performance comparison. It offers three computing paradigms: edge device on-device training, distributed computing, and single-machine simulation.

\makeatletter
\newcommand\justify{%
  \let\\\@centercr
  \rightskip\z@skip
  \leftskip\z@skip}
\makeatother
\begin{table*}[ht]
\centering
\begin{tabular}{||l | >{\justify}p{5cm}| >{\justify}p{5cm}||}
\hline
\textbf{Algorithm} & \textbf{Benefits} & \textbf{Drawbacks} \tabularnewline
\hline 
 \textbf{FedAvg} & - Simplicity \\
                   - Communication-efficiency \\
                   - Privacy-preserving 
                &  - Slow convergence speed \\
                   - Privacy risks \\
                   - Data heterogeneity 
      \tabularnewline
\hline
\textbf{FedSGD } &  - High convergence speed \\
                    - Handling data heterogeneity \\
                    - Localized Adaptation
                 &  - Hyper-parameter complexity \\
                    - Communication overhead \\
                    - Impact of imbalanced data 
      \tabularnewline
\hline
\textbf{FedProx } & - High convergence speed \\
                    - Handling non-IID data \\
                    - Regularization
                 &  - Hyper-parameter tuning \\
                    - Increased communication overhead \\
                    - Centralized aggregator
      \tabularnewline
\hline
\textbf{FedATT } &  - Fast learning  \\
                    - Communication efficiency  \\
                    - Handling non-IID data 
                 &  - Privacy risks \\
                    - Data heterogeneity \\
      \tabularnewline
\hline 
\textbf{SimFL } &   - Good accuracy \\
                    - Fast computation  \\
                    - Security      
                 &  - Communication overhead \\
                    - Impact of imbalanced data \\
      \tabularnewline
\hline
\end{tabular}
\vspace*{0.1cm}
\caption{FL algorithms: benefits and drawbacks.}
\label{table:3_1}
\end{table*}
In this section, we provided a taxonomy of the essentials of FL, including a concise definition of FL, its various categories, and the associated algorithms. Moving forward, we will delve into FL applications and explore how they are leveraged to enhance the capabilities of the 5G and 6G mobile generations.

\section{Federated Learning at PHY/MAC layers}
With the promising performance of ML in wireless communication \cite{qin2020federated}, there has been a surge in the area of ML for the wireless physical layer e.g., see \cite{qin2019deep} and \cite{o2017introduction}. FL is more communication efficient and privacy-preserving than traditional ML and, thus, it has a promising potential for addressing problems at the lowest layers of the OSI model \cite{elbir2021federated}. This section provides an overview of recent advancements in FL-based physical layer architecture, including symbol and signal detection, channel estimation, channel state information, beamforming, reconfigurable intelligent surfaces, channel access, and spectrum sharing.  
\subsection{Symbol/Signal Detection}
The receiver in a digital communication system must correctly identify the transmitted symbols from the observed channel output. This task is known as symbol detection \cite{farsad2020data}. The key benefit of using ML and DL techniques for symbol detection is to provide a data-driven mapping for modeling channel characteristics that model-based techniques do not effectively manage \cite{farsad2020datadriven}. In addition, since the model is directly fed with the obtained corrupted symbols, end-to-end learning allows the model to detect the symbols correctly without a channel estimation stage \cite{qin2019deep}. Using centralized learning for this role introduces high communication overhead and privacy issues, as well as limitations related to storing and transmitting a large amount of data. 

To overcome these challenges, the work in \cite{mashhadi2020fedrec} designed an FL-based receiver (FedRec) for symbol detection in downlink fading channels, designed to learn its mappings from a limited number of pilots. This approach takes advantage of this intuition by enabling users to collaborate for training through FL, which allows a single neural network to be trained over a diverse data set without additional pilots. Despite the need for numerous iterations of parameter exchanges with the base station (BS), FedRec yields an efficient symbol detector that achieves performance levels close to a MAP detector. This outperforms the model-based solution, especially when faced with inaccurate knowledge of the fading distribution. Furthermore, FedRec induces substantially less communication overhead than learning a neural network (NN) based symbol detector in a centralized scheme. FedRec facilitates utilizing compact NNs by avoiding feature extraction layers, which are trainable using relatively small datasets. While each local training data set encapsulates a relatively small number of fading channel realizations, the diversity among these data sets at different devices is exploited to obtain a unified model for all the devices by training in a federated manner.

The advantage of FL is recognizing signal modulation while protecting private data. As such, in \cite{shi2020signal}, instead of transmitting the raw data between the central server and all devices, only the convolutional NN (CNN) model updates are shared. The results show that the designed approach has reached more than 70\% recognition rate while satisfying privacy protection and data security.

The authors of \cite{modulationwang2021federated} presented an FL framework for automatic modulation classification named FedeAMC; under conditions of class imbalance and noise varying, data and training are in each local client, while only knowledge is shared with the server. The FL process contains six main steps: model and parameters choosing and broadcasting, local gradient calculation, key information uploading, key information aggregation, aggregated information downloading, and finally local model updating. The proposed approach achieves a low risk of data leakage without severe performance loss.
\subsection{Channel Estimation}
In wireless communication, there are some undesirable effects on the signals transmitted over the wireless channel caused by the physical properties of the channel. As a result, the signals arriving at the receiver end of the communication system are always attenuated, distorted, delayed, and phase-shifted \cite{oyerinde2012review}. Consequently, there is a need to provide a perfect and up-to-date estimation of the channel to compensate for these effects, and accurate signal demodulation, equalization, and decoding at the receiver end of the systems. Channel estimation through ML \cite{dong2019deep}  requires model training on a data set in a centralized manner, which typically includes received pilot signals as input and channel data as output \cite{elbir2020deep}. However, this method generates significant transmission overhead when collecting user data. To deal with this problem, FL was considered in \cite{elbir2020federated} to enhance the performance of channel estimation; instead of sending the entire dataset, only the model updates are transmitted between clients and server, which maintains good channel estimation and reduces the estimation error, as well as the transmission overhead that approximately 16 times lower than central learning. The authors designed a convolutional neural network at the BS and trained on the local data sets. The proposed approach has three stages: 
\begin{enumerate}
    \item \emph{Data collection:} Each user creates its local training dataset, which contains the input (signals pilot) and output (channel matrix) for model training.
    \item \emph{Training:} Each user works with its local dataset, computes model changes and sends them to the BS, where they are aggregated to train a global model.
    \item \emph{Prediction:} Each user will estimate its local channel by simply feeding the obtained pilot data into the trained model. 
\end{enumerate}
The fact that each client has access to the qualified model for channel estimation is a significant benefit of this method. 

The authors of \cite{xue2020low} used FL to create a channel prediction scheme to achieve channel pre-compensation for a low-cost free-space optical communication system. The proposed approach is effective in simplifying system structural and operational costs.
\subsection{Channel State Information}
To reap the full benefits of the reconfigurable intelligent surfaces (RIS) architecture. All efficient technologies must rely on the perfect channel state information (CSI) between the BS and the RIS and the perfect CSI between the RIS and users \cite{onggosanusi2018modular}. However, it is infeasible for RIS-enhanced systems to estimate correct CSI when radio frequency (RF) sensors or chains are not installed on the RIS. Therefore, employing FL for CSI detection in RIS-assisted wireless communications makes sense. In fact, for some small cellular BSs, a small amount of training data is not sufficient to produce a generalized model for CSI prediction. Centralized learning techniques combine all the data for processing and training, which results in communication overhead \cite{luo2018channel}. 

The authors in \cite{hou2021federated} introduced a decentralized approach to overcome these challenges. They proposed an FL-based framework for CSI prediction using three-dimensional convolutional neural networks. The global model is trained at macro BSs (MBs), by gathering all local datasets from the edge BSs. To deal with the performance gap between CL and FL, they designed the federated weights and gradients (FED-WG) algorithm, which works as follows:
\begin{enumerate}
    \item The MBs get the relevant local model weights and gradients from the BSs.
    \item FED-WG executes two consecutive rounds of parameter updates for the global model.
\end{enumerate}
The suggested FED-WG framework effectively narrows the performance gap between FL and CL, while significantly reducing the transmission overheads.

The general process of FL approaches typically assumes perfect knowledge of the CSI during the training phase, which is challenging for fast-fading channels. In addition, the literature uses a fixed number of clients to participate in the federated model's training. The work in \cite{pase2021convergence} tried to fill these gaps by proposing an FL approach where all clients use a constant global rate to complete training rounds. This consistently results in faster convergence, even when the CSI is imperfect.
\subsection{Beamforming}
Beamforming and massive multiple input multiple outputs (MIMO) systems aim to adapt the radiation pattern of an antenna array to particular scenarios. These technologies are a crucial part of 5G and 6G new radio (e.g., see \cite{yang2021federated} and \cite{dai2020deep}). Using a centralized learning scheme for hybrid beamforming has been widely studied. Incorporating FL into beamforming and massive MIMO systems holds the potential to enhance network efficiency, robustness, and scalability. As 5G and 6G technologies continue to evolve, integrating advanced learning techniques plays a pivotal role in shaping the future of wireless communication.

The authors in \cite{9177084} provide an FL-based framework to map the millimeter wave (mmWave) channels into analog beamformers in a multi-user downlink network. They built a CNN architecture at the BS, where the model is trained on the edge using only the gradients provided by the different users. The approach of \cite{9177084} proceeds as follows:
\begin{enumerate}
    \item The CNN model takes the channel matrix as input and the RF beamformer as output. The deep neural network is then trained by using the gradient data collected from the users. 
    \item Each user computes the gradient information with its available training data (a pair of channel matrices and corresponding beamformer index) and then sends it to the BS. 
    \item The BS receives all the gradient data from the users and performs the parameter updates for the CNN model which is comprised of 11 layers with two convolutional layers and a single fully connected layer. As a result, the beamformer indices are the CNN's output.
\end{enumerate}
FL strategy for hybrid beamforming provides more robust beamforming performance while exhibiting much less transmission overhead.

This framework requires the mmWave channel matrices as inputs, which is challenging to estimate and requires more significant training overheads. The authors in \cite{chafaa2021federated} suggested a beamforming scheme relies only on the channel state estimation at sub-6GHz, which is considerably more accessible with current technologies compared to mmWave channel estimation. The authors aim to predict the mmWave beamforming vectors exploiting sub-6GHz channels for a network composed of multiple access point–user links, where the deep neural network accepts the sub-6GHz channels as input and outputs directly the corresponding mmWave beamforming vector. Moreover, the authors suggest a distributed FL scheme to predict the beamforming vectors locally at each user without uploading the local data to a central server. There are three main benefits of this framework:
\begin{itemize}
    \item The computation load is distributed to the edge of the network as opposed to a centralized cloud-based approach;
    \item Users only share their neural network parameters and not their data, which highly reduces the signaling load, and protects the data of the individual users; 
    \item Users still share their acquired knowledge of the environment to improve the quality of their predictions.
\end{itemize}

mmWave-MIMO communication systems substantially improve the throughput in 5G networks \cite{heath2016overview}. However, these solutions reduce the multiplexing gain in the case of a small number of RF chains. The work in \cite{elbir2021federated7} introduced spatial path index modulation (SPIM), a technique for improving gain using other signal bits modulated by spatial path indices. The authors proposed model-based and model-free frameworks for beamformer design in multi-user SPIM-MIMO systems; they first designed the beamformers via a model-based manifold optimization algorithm. Then, the authors train a CNN model on the local dataset using FL and dropout learning to mitigate overfitting. The model updates are then gathered at the BS for aggregation and then sent back to the users for the prediction phase, where the model estimates beamformers by feeding them their channel data. The advantages of utilizing dropout learning in FL include:
\begin{enumerate}[label=\Alph*]
    \item Achieving approximately 10 times lower overhead than centralized learning. 
    \item Reducing the communication cost during training.
    \item Achieving superior spectral efficiency compared to model-based SPIM and conventional mmWave-MIMO.
\end{enumerate}
\subsection{Reconfigurable Intelligent Surfaces}
Reconfigurable Intelligent Surfaces (RIS) have emerged as a strong contender in comparison to massive MIMO. This technology has garnered notable research interest \cite{saad2019vision}, particularly in harnessing recent Machine Learning techniques, especially FL, to enable reconfigurable propagation settings in wireless communication capabilities. The integration of RIS enhances the beamforming gain in uplink communications within a massive MIMO system \cite{wang2020intelligent}. Nonetheless, this approach introduces several challenges such as ignoring user privacy and producing high communication overhead.

To overcome these challenges, the study outlined in \cite{ma2020distributed} introduced an FL-driven beamformer design tailored for an IRS-assisted scenario. This entails an algorithm referred to as "optimal beam reflection based on FL" aiming to enhance high-speed communication with sparse channel state information (CSI) by improving the data rate and protecting privacy. The standard FL algorithm is adopted to train a regression multi-layer perceptron with model transmission; the proposed framework effectively reaches a theoretical value that exceeds 90\% of what is achieved by centralized Machine Learning, all while ensuring user privacy protection. Nevertheless, the RIS-user link is the only channel that is designed, while the BS-RIS link is supposed to be constant. However, in a practical case, the mm-Wave channel tends to be highly complex and characterized by a limited channel length, largely owing to environmental fluctuations. Hence, the solution of \cite{elbir2020federated} estimates both direct (BS-user) and cascaded (BS-IRS-user) channels in an RIS-assisted scenario, where input and output data are combined for each communication link only a single CNN architecture is designed instead of different NNs.

A privacy-preserving paradigm combining FL with RIS in the mmWave communication systems is designed in \cite{li2020enhanced}. First, the private data handled on each local device trains and encrypts the local models. Then, at the central server, a global model is created by aggregating them. While ensuring user data confidentiality, the suggested technique successfully approaches theoretical value and achieves greater than 95\% of that created by centralized ML.
\subsection{Channel Access}
ML and DL techniques have significantly enhanced the design of the MAC layer and have also contributed to the improvement of channel access performance by automating the tuning of protocol parameters and the evaluation of networking protocols. These advancements address critical issues related to the limited power of wireless devices and privacy concerns \cite{liu2006rl,qiao2018intelligent,pasandi2020mac}. FL has garnered considerable interest due to its inherent parallelization capabilities and its potential for greater efficiency compared to centralized methods in terms of storage and privacy. 

The authors in \cite{amiri2020machine} proposed an alternative analog communication approach in which the devices transmit their local gradient estimates directly over the wireless channel. The learning process is motivated by an FL strategy where devices have their local datasets. They communicate with the parameter server over a wireless MAC. Here, the parameter server is not interested in the individual gradient vectors but in their average, and the wireless MAC automatically provides the server with the sum of the gradients. 

Non-orthogonal multiple access (NOMA) technology is considered a building block for 5G and beyond networks. The study discussed in \cite{habachi2019fast} investigates the organization of machine-type devices (MTD) into clusters for NOMA-based systems. Focusing on resource allocation, the authors put forth an approach involving traffic model estimation through FL. In this method, the MTD autonomously estimates its traffic model parameters and transmits them to the BS, which subsequently aggregates the traffic model, allocates the suitable resource blocks, and transmits power to each MTD.

The authors in \cite{zhong2021mobile} proposed a NOMA-enhanced wireless network model to offer NOMA desired channel conditions and increase consumer quality by adopting the FL concept to enable multiple agents to simultaneously explore similar environments and exchange experiences.
{\renewcommand{\arraystretch}{1.5}%
\begin{table*}[ht!]
\centering
\begin{tabular}{||l|l|p{5cm}|p{5cm}||}
 \hline
 \textbf{Ref} & \textbf{Application} & \textbf{Benefits} & \textbf{Drawbacks} \\ [0.5ex] 
 \hline
 \cite{mashhadi2020fedrec,shi2020signal,modulationwang2021federated}
 & Symbol / Signal Detection &  Learning collaboratively to clear the received symbols, without the need for channel estimation stage or gathering raw data & Need deep ML models for better performance \\
 \hline
 \cite{elbir2020federated,xue2020low} & Channel Estimation & Ability of each user to estimate its channel with less transmission overhead & Heavy computation resources
 due to the labeling phase of the channel data \\
 \hline
 \cite{9177084,chafaa2021federated,elbir2021federated7}
 & Beamforming &  Dynamic beamforming and antenna configurations for optimal signal transmission, improving coverage and reducing interference & Sub-optimum performance and complex labeling\\
 \hline
 \cite{ma2020distributed,elbir2020federated,li2020enhanced}
 & RIS &  Achieving high speed convergence in high dimension and complex environment & Only RIS beamforming is performed \\
 \hline
 \cite{hou2021federated,pase2021convergence} & CSI prediction &  
 Estimating the accurate CSI when the RF chains or sensors are not equipped on the RIS while preserving users' privacy & Need to maximize the training performance and the heterogeneous data should be taken into consideration \\ 
 \hline
 \cite{amiri2020machine,habachi2019fast,zhong2021mobile} & 
 Channel Access & Tuning automatically individual protocol parameters while improving the convergence speed and communication delay & The impact of CSI estimation error on the performance needs to be studied.\\
 \hline
 \cite{troglia2019fair,wang2020privacy} & Spectrum Sharing &  
 Improving spectrum sharing efficiency by learning in a distributed way from local information, without requiring the entire data & Need to collect more real data to analyze the true potential of FL in spectrum sharing\\
 \hline
\end{tabular}
\vspace*{0.1cm}
\caption{Summary of FL-based contributions at PHY/MAC layers.}
\label{table:4}
\end{table*}}
\subsection{Spectrum Sharing}
Spectrum efficiency is one of the key performance metrics in 5G communication networks and even beyond \cite{zhang2017survey}. Advanced spectrum-sharing techniques, such as ML and DL \cite{liang2019spectrum}, are generally used to deal with typical challenges while enhancing efficiency. However, deploying ML techniques for spectrum applications faces the challenges of limited dataset resources, as well as privacy concerns. Thus, FL scheme handles privacy challenges while achieving interesting results compared to centralized learning.

In \cite{troglia2019fair}, the authors proposed an ML-based non-coherent spectrum sensing system named FaIR, which leverages a communication-efficient distributed learning framework, FL, for environmental sensors to collaborate and train a data-driven ML model for incumbent detection under minimal communication bandwidth. The preliminary results show that state-of-the-art spectrum classification algorithms in an FL environment perform better than traditional techniques.

In \cite{wang2020privacy}, the authors investigated the spectrum sharing problem in high-mobility vehicular networks by adopting the FTL method. They start by modeling resource sharing as a multi-agent reinforcement learning problem, which is then solved using a fingerprint-based deep Q-network method that is amenable to a distributed implementation. The results show that with a well-designed reward system and training process, the vehicle-to-vehicle transmitters learn from their interactions with the communication environment. The authors developed an effective strategy to collaborate in a distributed manner, thereby optimizing system-level performance using local information. 
\subsection{Summary}
In this section, we have investigated the existing FL-based schemes that improve communication quality and key performance indicators (KPIs) for physical and MAC layers in wireless communication systems. 
The following are the lessons learned from this section :
\begin{itemize}
    \item The integration of FL at the PHY and the MAC layers leverages the unique capabilities of FL to optimize wireless communication processes, enhancing efficiency, reliability, and performance.
    \item FL represents a novel approach that redefines how wireless networks operate at the most fundamental level such as signal detection, channel estimation, beamforming, RIS optimization, channel access, and spectrum sharing while presenting certain drawbacks such as model/data/hardware complexity, communication efficiency, and learning accuracy.
\end{itemize}
Table \ref{table:4} illustrates a summary of the main contributions and their respective benefits and drawbacks.

\section{Federated Learning at NET Layer}
The use of FL in the network layer offers a privacy-preserving, efficient, and scalable approach to ML. It addresses the challenges posed by distributed and sensitive data, while also catering to the low-latency and fault-tolerant requirements of networked systems. It is applied in several tasks, such as resource management, user behavior prediction, network slicing, routing optimization, and cloud computing, among others, as detailed next. 
\subsection{Traffic Classification}
Traffic classification (TC) has received increasing attention in recent years. It aims to offer the ability to automatically recognize the application that has generated a given stream of packets from the direct and passive observation of the individual packets flowing in the network \cite{valenti2013reviewing}. Due to the achievement of FL in data privacy protection, it is applied even in internet traffic classification to improve the accuracy of packet transmission while preserving private data. 

The work \cite{mun2021internet} proposed an FL-based traffic classification framework named FLIC that classifies new applications on the fly when a client joins the learning process with a new application. The designed protocol achieves an accuracy comparable to the centralized scheme for identifying internet applications with a privacy guarantee. 

The work in \cite{majeed2020cross} built a cross-silo horizontal federated model for TC using flow-based time-related features. The proposed FL framework performs comparably to a centralized DL model for internet application identification without privacy leakage.

In \cite{peng2021federated}, a novel approach using federated semi-supervised learning for network traffic classification was designed. The federated servers and several clients work together to train a global classification model, where unlabeled data is used on the client, and labeled data is used on the server. Results show that this method protects users' privacy without sharing a large amount of labeled data as well as achieving good accuracy.

\subsection{Resource Management}
Power and resource allocation remain critical wireless challenges due to the need for more resources \cite{nazir2021power}. Optimizing a multi-cell network spectral efficiency and connectivity invariably results in non-convex resource allocation issues, which are frequently handled using basic methods such as successive convex approximation and matching theory. To address the high complexity and impracticality of traditional approaches, the authors in \cite{ale2019online} and \cite{dong2019deep2} used ML algorithms to solve resource optimization of problems related to mobile edge computing (MEC). However, sending all local information to a central controller in MEC-enabled high-altitude balloon (HAB) networks is impracticable since the transmission of local datasets results in substantial energy usage. To this end, FL allows dispersed devices to cooperatively train an ML model by sharing trained parameters with other devices instead of sending a huge dataset.

The authors in \cite{wang2021federated} solved this problem using a support vector machine (SVM) based FL to determine the user association proactively. The proposed method enables each HAB to cooperatively build an SVM model to determine all user associations without transmitting computational tasks or user historical associations to other HABs. The SVM model analyzes the relationship between the future user association and the data size of the task that each user needs to process at the current time slot to determine the user association proactively. Given the prediction of the optimal user association, each user's service sequence and task allocation are optimized to minimize the weighted sum of the energy and time consumption. This approach reduces the weighted sum of the energy and time consumption of all users by up to 16.1\% compared to a conventional centralized method.

In \cite{yan2020federated}, the authors investigated distributed power allocation for edge users in decentralized wireless networks. The proposed model (FL-CA) aims to minimize power consumption while satisfying the user quality of service (QoS) requirement and protecting user privacy. In the FL framework, edge devices locally make decisions on power allocation by training a local actor-critic model and then send the gradients and weight ages generated by the actor-network to the BS for information aggregation at regular intervals. Moreover, the authors adopt the federated augmentation algorithm which uses the Wasserstein generative adversarial network (WGAN) for data augmentation to overcome the over-fitting problem caused by data leakages. The algorithm empowers each device to replenish the data buffer using a generative model of WGANs until reaching an i.i.d training dataset, which reduces the communication overhead compared to direct data sample exchanges.

In \cite{ali2021federated}, the authors proposed a federated reinforcement learning (FRL) based channel resource allocation framework for ultra-dense 5G and B5G wireless networks and suggested collaborative learning estimates for faster learning convergence. The results demonstrate that the designed FRL model is superior to non-federated reinforcement learning. 
\subsection{Radio Access Network}
The radio access network (RAN) is a fundamental part of a mobile telecommunication system where user data is sensitive and subject to strict privacy regulations. Centralized ML approaches raise privacy concerns due to data aggregation. FL allows local model training on devices without sharing raw data, preserving user privacy while enabling collective model updates \cite{foukalas2022federated,lopez2022survey,mahrez2023benchmarking}.

Deep reinforcement learning was discussed in \cite{abouaomar2022federated} and associated challenges in a multi-mobile virtual network operators (MVNOs) environment. The authors designed a federated deep reinforcement learning mechanism on an O-RAN architecture to improve the radio resource allocation operation of MVNOs and illustrate through extensive simulations that the proposed RAN slicing mechanism enables a better allocation of the needed radio resources to satisfy users’ QoS requirements in terms of delay and data rate.

The work in \cite{manzoor2022federated} proposed a novel FL-based mobility and demand-aware proactive content offloading (MDPCO) framework that exploits distributed learning strategies and capitalizes on users’ mobility and demand information for proactive content offloading. The efficacy of MDCPO improves performance against local and cloud-based models.

The authors in \cite{zhang2022federated} designed a federated deep reinforcement learning algorithm to coordinate multiple independent xAPPs in O-RAN for network slicing by developing two xAPPs, namely a power control xAPP and a slice-based resource allocation xAPP to enhance learning efficiency and improve network performance. 
\subsection{User Mobility/Location Prediction}
Due to the varied quality-of-service requirements, user behavior and wireless network performance have become crucial for developing and evaluating new applications and service opportunities. FL strategy is promising in several situations to predict users' behavior and maximize the quality of experience (QoE). Based on the mobility predictions, we provide more information about the network, the users dynamically choose a subchannel to upload data in the uplink, the BS dynamically allocates multiple subchannels to multiple users in the downlink, and multiple users who occupy the same subchannel perform NOMA or full-duplex. Due to the privacy-sensitive nature of user activities, existing location prediction approaches \cite{luca2020deep,feng2018deepmove,anagnostopoulos2011mobility} rely on centralized storage of user mobility data for model training, which raises privacy concerns and risks. FL has the potential to enable predictive features on smartphones without diminishing the QoE or leaking private information. 

The authors in \cite{li2020predicting} presented an adaptive FL model as a decentralized method for mobility prediction. They combined a personalized FL model with an attention network to predict user location in a protected way. This approach leverages the useful information in the behaviors of massive users to train accurate mobility prediction models and, meanwhile, remove the need for centralized storage.

Another interesting work on human trajectory prediction using FL is presented in \cite{feng2020pmf}. The authors proposed a privacy-preserving human mobility prediction framework to achieve promising prediction performance while preserving personal data on local devices. Based on the DL mobility model, no private data is uploaded to the centralized server. The only uploaded thing is the updated model parameters which are difficult to crack and thus more secure.

The authors in \cite{sozinov2018human} demonstrated that FL for human activity recognition tasks produces models with slightly worse, but acceptable accuracy than centralized techniques. Recently, it has been used in several works and in proactive HO mm-wave vehicular networks to protect user position information, reduce communication overhead, and reduce frequency. The authors of \cite{wu2021hierarchical} proposed a hierarchical personalized FL (HPFL) approach as a novel client-server architecture framework to serve FL in user modeling with inconsistent clients. Traditional FL indiscriminately aggregates and updates the consistent whole user models, but HPFL attempts to divide and process the different components of the heterogeneous models independently.

Location information is a service enabler for communication network design, operations, and optimization \cite{bourdoux20206g}. FedLoc is a new collaborative positioning and location data processing framework proposed in \cite{yin2020fedloc}. This approach properly resolves the question of privacy in target localization and location data processing in collaboration with many mobile users. The federated approach is promising thanks to its significant advantages:
\begin{itemize}
    \item Efficiency in handling the data privacy issue, allowing mobile users to exchange location-related information safely.
    \item Collaboration among mobile users facilitates the calibration effort. 
    \item Cell phones are becoming capable platforms for performing complex calculations.
\end{itemize}

The work in \cite{ciftler2020federated} presented another FL-based approach to improve the accuracy of received signal strength fingerprint-based localization while protecting the privacy of the crowdsourcing participants. The main idea of this novel method is to keep local data where it is generated and convey only local models throughout the learning process. When employed as a booster for centralized learning, the suggested approach enhanced localization accuracy by 1.8 meters and reached satisfactory localization accuracy when used alone.

The authors of \cite{xiao2021federated} used FL for human activity recognition (HAR) allowing each user to handle his activity recognition task safely and collectively. They designed a perceptive extraction network as the feature extractor for each user to capture enough features from HAR data. PEN comprised of a featured network to discover local features and a relation network, a combination of long short-term memory (LSTM) and attention mechanism, which is in charge of mining global relationships hidden in the data.   
\subsection{Network Slicing}
Network slicing enables the shift from a network as an infrastructure configuration to a network as a service to support a wide range of 5G and 6G smart services with several requirements \cite{khan2020network} \cite{kazmi2019network}. By integrating FL into network slicing, telecommunication providers and service operators unlock the potential of personalized services while maintaining data privacy and network performance. As the demand for customizable and secure services grows \cite{wu2021ai}, the combination of FL and network slicing is likely to play a significant role in the future of telecommunications \cite{khan2020federated}. 

The authors in \cite{wang2019edge} proposed integrating an FL framework with mobile edge systems to train double-deep Q-learning agents at the network edge for caching and computational offloading decisions while protecting user privacy. The “In-Edge AI” was assessed and shown the capacity to achieve near-optimal performance with a reasonably minimal learning overhead, while the system is cognitive and adaptive to mobile communication networks.

In \cite{messaoud2020deep}, the authors developed an FL-based approach to enhance resource allocation strategy in multi-industrial IoT. This novel proposal deep federated Q-learning, involves IoT slices' resource allocation in terms of transmission power and spreading factor according to the slices QoS' requirements.

The authors of \cite{brik2020predicting} adopted FL to predict slices’ service-oriented KPIs by keeping raw data where it is generated and sending only users’ local models to the centralized entity for aggregation. The obtained results showed the efficiency of the designed approach in reaching good prediction accuracy while ensuring privacy issues.

Aiming to improve network throughput while reducing hand-off cost, the authors of \cite{liu2020device} designed an efficient device association scheme for RAN slicing by exploiting a hybrid FRL framework. Two levels of aggregation were proposed for the device association problem. One involves the same type of services to aggregate the local parameter models to share similar samples. The other one pertains to the different types of services to aggregate access features to make an optimal global decision on network slicing and BS selection.

The authors of \cite{li2020federatedslicing} investigated the distributed network slicing for 5G. They introduced a novel framework centered around a federated orchestrator in the control plane. This entity is responsible for coordinating spectrum and computational resources, all without requiring any exchange of personal data or resource information from the local BS. This framework significantly reduces service response time for both supported services, especially compared to network slicing with only a single resource.
\subsection{Routing Optimization}
Predictive routing employs the power of AI to constantly evaluate real-time data to anticipate outcomes \cite{sendra2017including}. Most existing ML techniques to drive routing and traffic control choices are essential \cite{yao2019ai}. However, they rarely focus on challenging networks, characterized by constantly changing network conditions and a significant volume of traffic generated by edge devices. Several FL systems are proposed to address those problems and optimize routing decisions in challenging scenarios.

To increase performance and facilitate the scalability of data-intensive applications, the authors in \cite{sacco2020federated} designed a federated architecture for routing packets inside a distributed edge network. They proposed a novel path selection model to predict the best route using LSTM. When a peak load is anticipated, the controller leverages this information to make adjustments to routing. By employing FL, all controllers attain a global view of the infrastructure while exchanging relatively little information, which helps organizations protect the bandwidth for the application traffic.

The authors of \cite{cao2021mobility} developed mobility-aware routing and caching strategies
for dense small-cell networks based on the FL framework to optimize the cache placement and minimize the network cost. They first started by segmenting the whole geographical region into small sections, each with one small BS (SBS) and many mobilized users (MUs). They proposed a federated routing and popularity learning strategy in which the SBSs jointly learn the routing and preferences of their respective MUs and make caching decisions.

In \cite{wilbur2020time}, the authors created a decentralized routing application on time-dependent transportation networks using FL to learn shared prediction models online. All routing occurs at the network edge within a private roadside unit fog network, where access to outside cloud services is assumed to be intermittent. This system avoids the costly transfer of raw data between processes.
{\renewcommand{\arraystretch}{1.5}%
\begin{table*}[htp!]
\centering
\begin{tabular}{||p{3cm}|p{3cm}|p{5cm}|p{5cm}||}
 \hline
 \textbf{Ref} & \textbf{Application} &  \textbf{Benefits} & \textbf{Drawbacks} \\ [0.5ex] 
 \hline
 \cite{mun2021internet,majeed2020cross,majeed2021cross,peng2021federated} & Traffic Classification & Safely classifying internet traffic as well improving the accuracy of packet transmission & More improvements are needed to deal with complex data and information security \\ 
 \hline
 \cite{wang2021federated,yan2020federated,ali2021federated}
 & Resource management & FL facilitates adaptive resource allocation, distributing bandwidth, power, and other resources intelligently across network nodes & Need more evaluations by considering for practical applications \\
 \hline
  \cite{abouaomar2022federated,manzoor2022federated,zhang2022federated} 
 & Radio access network & Satisfy users’ QoS requirements in terms of delay and data rate & Need more enhancement privacy beyond what FL inherently promotes\\
 \hline
\cite{li2020predicting,feng2020pmf,sozinov2018human,wu2021hierarchical,yin2020fedloc,ciftler2020federated,xiao2021federated} & User mobility/location prediction &  Enabling predictive features on smartphones without diminishing the QoE or leaking private information & Mobile terminals should be able to process a modest amount of data and perform analysis with FL frameworks\\
 \hline
 \cite{wang2019edge,messaoud2020deep,brik2020predicting,liu2020device,li2020federatedslicing} & Network slicing &  Offering significant improvement on service latency performance for all supported services & Updating local models with high frequency increases the network overhead and consumes network and computing resources \\
 \hline
 \cite{sacco2020federated,cao2021mobility,wilbur2020time}
 & Routing optimization & Optimizing routing for challenging scenarios while avoiding costly transfer of raw traffic data and reducing bandwidth stress & Need more travel time improvement and grid prediction models to mitigate the impact of errors on user trips \\
 \hline
 \cite{YAZDINEJAD2021102574,lim2021towards,wang2021uav,wang2020learning,zeng2020federated,zhang2020federated,mowla2019federated} & UAV flying networks & Handling target challenges in UAV systems as well as operating at a considerably lower communication cost & Need to accelerate the strategy-making process and the resource allocation should be optimized \\ 
 \hline
  \cite{kwon2020multiagent,zhao2021federated} & Underwater Networks
  & Enhancing underwater communication systems performance with high transmit power and secure underwater scenarios & The learning process should be more accelerated and applied in real scenarios \\ 
 \hline
\cite{fang2020highly,zhang2020efficient,liu2020client,zhao2020federated,wu2020content,qu2020decentralized} & Cloud/Fog computing 
 & Achieving less energy consumption to make full use of the big data in both cloud and fog computing & The training efficiency has to be enhanced as well as optimizing some critical parameters
\\
 \hline
\end{tabular}
\vspace*{0.1cm}
\caption{Summary of FL-based contributions at NET layer.}
\label{table:5}
\end{table*}}
\subsection{Unmanned Aerial Vehicle}
The combination of AI and Unmanned Aerial Vehicles (UAVs), commonly known as drones, holds significant promise for various industries and applications \cite{khuwaja2018survey,hassija2021fast,bithas2019survey}. However, it presents challenges related to regulation, safety, privacy, security, and complexity. FL is an emerging and promising application that combines privacy preservation, low latency, low communication overhead, and real-time learning with the capabilities of UAVs \cite{brik2020federated}.

The work in \cite{YAZDINEJAD2021102574} presented an FL-based drone authentication model with drones’ RF features in IoT networks. The deep NN (DNN) architecture is used, with stochastic gradient descent optimization performed locally on the drones. Results show that the federated drone authentication model gains a high true positive rate during drone authentication and better performance compared to other ML systems.

The study in \cite{lim2021towards} adopted FL to facilitate privacy-preserving sensing and collaborative learning in the UAV services. This work proposed a multi-dimensional contract matching design that aims to match the most optimal UAV to each sensing sub-region while accounting for UAV heterogeneity.

The authors of \cite{wang2021uav} proposed a hierarchical nested personalized FL, a holistic distributed ML framework for personalized model training across the worker leader core network hierarchy. This method used data sharing across device clusters to create tailored local models.

In \cite{wang2020learning}, a secured FL framework for UAV-assisted MCS was presented. Firstly, three attacks were proposed, and the related countermeasures were studied to secure cooperative learning for UAVs. Secondly, an FL approach has been developed using the blockchain network to securely store and trace UAV contributions in an immutable way while protecting local model exchange across UAVs. Finally, a privacy-preserving local model-sharing mechanism was developed to protect UAVs with high aggregation accuracy.

The authors of \cite{zeng2020federated} proposed a novel architecture for implementing distributed learning algorithms inside a UAV swarm that consists of a leading UAV and several following UAVs. Each UAV trains a local FL model based on its gathered data and then sends it to the leading UAV, aggregating the received models, generating a global federated model, and transmitting the knowledge to followers through the intra-swarm network.

In \cite{zhang2020federated}, the authors proposed FL-aided multi-UAV systems for task classification in the broad application area of exploration scenarios. At first, local training is carried out by each UAV relying on its own locally gathered images; the model developed from the local training is then transmitted to the GFC over fading wireless channels. Finally, a global model is built at the GFC and sent back to each UAV for the next local model update. This framework operates at a considerably lower communication cost compared to traditional ML.

FL is a promising solution for flying ad-hoc networks (FANET). The authors of \cite{mowla2019federated} designed an FL-based on-device jamming attack detection security architecture for FANET. They added a client group prioritization technique leveraging the Dempster–Shafer theory \cite{sentz2002combination}, which allows the aggregator node to identify better client groups for calculating the global update.
\subsection{Underwater Networks}
Underwater wireless communication is another type of wireless communication in which acoustic signals across an underwater channel send digital information. Numerous efforts have been made to enhance the performance of underwater communication systems using ML, DL, and transfer learning techniques. However, the set-up of underwater networks is challenging because of the high propagation delay and high transmit power compared to wireless free-space air networks \cite{domingo2011securing}. FL mechanism holds promise as a secure means to collectively train underwater communication models \cite{victor2022federated}.

The authors of \cite{kwon2020multiagent} proposed a novel multi-agent deep reinforcement learning algorithm to remove unexpected smart ocean environment changes and channel unreliability. First, each device conducts distributed DL training with personal data, aggregating the output at a centralized station. Then, the centralized machine computes global optimal solutions and then distributes the results to the associated participants. This approach effectively enhances the throughput performance in underwater scenarios.

For the ocean of things, the authors in \cite{zhao2021federated} developed a federated meta-learning enhanced acoustic radio cooperative framework that takes advantage of data distributed across surface nodes to train the DL-receiver in the context of random scheduling wireless networks. This approach enables distributed transfer learning to be adapted to new datasets. 
\subsection{Cloud/Fog Computing}
While the cloud proves a powerful computing platform in numerous applications, it encounters difficulties in scenarios with rigorous requirements including drones and autonomous vehicles that need low-latency and privacy-preserving \cite{zhu2020toward}. FL is promising for training ML models on mobile devices while ensuring limited computing, storage, energy, and bandwidth.

In \cite{fang2020highly}, the authors presented a highly efficient FL framework with strong privacy preservation in cloud computing. Instead of noise injection methods, a lightweight encryption protocol is employed, giving both provable privacy and desirable model utility. This approach provides secure training to learn from all available data. 

Existing FL approaches for cloud-based AI IoT (AIoT) applications need more accurate prediction results. The authors in \cite{zhang2020efficient} presented a collaborative architecture that deploys DNN models across both cloud and AIoT devices. They designed two-stage training and co-inference schemes to improve the prediction accuracy of individual AIoT devices using little branches, reducing their average inference time with the help of the big branch part deployed on the cloud.

A client-edge-cloud hierarchical FL system was presented in \cite{liu2020client} based on the HierFAVG algorithm that allows multiple edge servers to perform partial model aggregation. This framework simultaneously reduces the model training time and the energy consumption of the end devices compared to traditional cloud-based FL. 

To take full advantage of fog computing and AI \cite{habibi2020fog}, a new paradigm of FL-enabled intelligent fog radio access networks was proposed in \cite{zhao2020federated}. This framework supports intelligent signal processing and network management with low communication cost and high efficiency, providing a feasible approach to implementing network edge intelligence.

The authors in \cite{wu2020content} used FL to predict content popularity in fog radio access networks and reduce computational complexity. The inputs of the prediction model are the averaged popularity scores for clustered users' contents, the model is then learned automatically by training the algorithm with historical popularity and the pre-processed inputs. 

To deal with the privacy gap in fog computing, the authors of \cite{qu2020decentralized} created a blockchain FL system that enables autonomous ML without any centralized authority to maintain the global model and coordinates using blockchain. The latency performance derives the optimal block generation rate by considering communication, consensus delays, and computation costs.

\subsection{Summary}
This section covers recent findings on applying FL mechanisms at the network layer. This emerging paradigm effectively approaches the results obtained by centralized ML techniques while guaranteeing 5G and 6G requirements such as user privacy, latency, limited computing, energy, and bandwidth. The lessons learned from this section are as follows:
\begin{itemize}
    \item The application of FL at the network layer redefines how networks are managed, optimized, and evolved. By leveraging FL's decentralized learning approach, the network layer gains the ability to enhance its performance, responsiveness, and intelligence while addressing critical challenges (e.g., privacy).
    \item While FL applications at the network layer promise numerous benefits, they also present some challenges including communication overhead, synchronization, security, and scalability. 
\end{itemize}
Table \ref{table:5} illustrates a summary of FL-based contributions that enhance the network layer architecture.

\section{Federated Learning at APP/Transport Layers}
The application and the transport layers are the last components of the OSI model that specify the shared protocols and interface methods used by hosts in a communications network. FL significantly improves handling APP layers challenges in wireless communication networks. In this section, we provide a deep dive into recent FL contributions in the context of APP and transport layers. 
\subsection{Transport Performance}
FL is applied to enhance the performance and efficiency of the data transport protocol and the transmission control protocol (TCP) used for reliable data transmission over the Internet. It brings several benefits, including improved congestion control, network management, and protocol optimization \cite{renda2022federated}. 

To enhance the performance of the internet of vehicles, the authors in \cite{pokhrel2020improving} proposed a privacy-preserving FL framework that guarantees ubiquitous connection among the vehicles and presented a local mechanism to satisfy the reliability and latency requirements. This work introduced a mathematical framework that accurately estimates TCP-perceived packet loss probability and latency for vehicles when wireless channels suffer from different magnitude losses and transmission delays.

The authors of \cite{pokhrel2020compound} adopted FL to study the performance of long-lived compound TCP (C-TCP) flows over Industry 4.0 WiFi infrastructure. They developed a novel analytical model to investigate the impacts of collisions, wireless losses, and AP buffer losses on C-TCP over WiFi with bidirectional flows. The results demonstrate that cognitive radio and FL significantly enhance the performance of the industrial multiple scenarios.
\subsection{Cooperative and Distributed Computing}
Distributed computing is a paradigm where elements of a software system are distributed across multiple computers, yet function collectively as a unified system. This strategy is employed to enhance efficiency and performance. Combining AI and distributed computing poses challenges in data management and communication between nodes. FL paradigm is used to safely manage distributed computing systems by training a shared model collaboratively while keeping all data on their devices. In \cite{sharma2020blockchain}, the authors designed a distributed computing defense framework for a sustainable society by combining FL and blockchain to preserve privacy and unlock the full potential of ML in distributed computing environments. Compared to traditional approaches, the result outcomes are promising in terms of accuracy and loss.

Recently, cooperative computing has emerged as a novel paradigm within the domain of distributed strategies. This approach revolves around fostering collaboration and synergy among various computing nodes to achieve enhanced system performance and efficiency. It capitalizes on the collective intelligence and resources of individual components, paving the way for innovative solutions that address complex challenges. In \cite{savazzi2020federated}, the authors introduced a novel FL framework that leverages the cooperation of devices that perform data operations inside the network by iterating local computations and mutual interactions via consensus-based methods. This method paves the way for FL integration in 5G and beyond, characterized by decentralized connectivity and computing.
\subsection{Crowdsourcing / Crowdsensing}
Crowdsourcing has the potential to be highly advantageous to accomplish computing tasks due to its ability to rapidly collect information \cite{wazny2017crowdsourcing}. FL is used to safely implement crowd computing. It involves training ML models using decentralized data from multiple servers, without transferring the data to a central location. This helps to protect sensitive data while sharing learning and analysis.

The authors of \cite{li2020crowdsfl} presented a crowdsourcing framework based on FL, blockchain, and re-encryption technology. All blockchain nodes participate in crowdsourcing to retain a full data backup. By adopting FL, each worker uploads the model or gradient data to clients for aggregation. This approach allows users to implement decentralized crowdsourcing with less overhead and higher security while improving the computing audit ability.

The work in \cite{zhao2020local} studied FL and local differential privacy to facilitate the crowdsourcing use cases. This approach enables vehicular crowdsourcing applications to train ML models and predict traffic status while preserving privacy and reducing communication costs.

Recently, crowdsensing has gained significant attention and becoming an appealing paradigm for sensor-based data collection \cite{capponi2019survey}. However, the diverse and sensitive nature of data collected through crowdsensing necessitates robust privacy protection mechanisms to secure individual user identities and personal information. Addressing this challenge is imperative to unlock the full potential of crowdsensing while upholding privacy standards. Several recent contributions employed FL paradigms to overcome those limitations. In \cite{wang2020federated}, a federated crowdsensing framework was proposed to analyze the privacy concerns of the four crowdsensing stages: creation task, assignment task, execution task, and data aggregation task. 

The authors of \cite{zhao2021crowdsensing} integrated FL into mobile crowdsourcing to design a privacy-preserving system named PriFedAvg where the participants locally process sensing data via FL and protect training by only uploading the encrypted training models.

The quality-aware user recruitment problem was investigated in \cite{zhang2021quality}, using FL to predict the quality of sensed data from different users by analyzing the correlation between data and context information through FL. This method improves the quality of sensed compared with traditional algorithms.

A novel approach was developed in \cite{chen2021federated} for fake task detection based on horizontal FL. Two ML algorithms and datasets were implemented to identify fake tasks containing several independent detection devices and an aggregation entity.
\subsection{Quality of Experience}
The evolution of streaming services needs high data rates, low latency, and good QoS. Several challenges must be addressed while delivering a specific service including the large number of devices, the heterogeneous networks, and the uncontrollable environments. The quality of experience (QoE) concept is gaining tremendous research efforts to improve and deliver reliable and added-value services, at a high user experience \cite{bouraqia2020quality}. ML-based QoE models suffer from overfitting due to low data volume and limited participant profiles \cite{aroussi2014survey}. A privacy-preserving ML scheme such as FL, enables the sharing of QoE data models between all participants only by transmitting model parameters.

The authors in \cite{ickin2019privacy} adopted FL and round-robin learning, where the model is trained sequentially amongst the collaborated partner nodes to show that high accuracy is achieved without sharing sensitive data between participants.

To facilitate smooth QoE model management and obtain higher accuracy, the work in \cite{ickin2020ensemble} presented a set-based Bayesian synthetic data generation method for FL that effectively reduces the training time by 30\% and the network footprint in the communication channel by 60\%.
\subsection{Cybersecurity}
Cybersecurity benefits from AI techniques to enhance security performance and better protection against complex threats \cite{wirkuttis2017artificial}. However, traditional ML architecture does not meet the demand of cybersecurity because of the large amount of data, the data heterogeneity, the high data velocity, and the major existing challenges in gathering cyber intelligence and attack datasets from distributed resources. The FL paradigm is used to securely train models for anomaly detection tasks without collecting raw data \cite{rey2022federated}.

A privacy-preserving approach was proposed in \cite{xu2019verifynet} to support verification of the server’s calculation results to each user which allows all participants to verify the results' correctness returned from the central server with acceptable overhead.

The authors of \cite{khramtsova2020federated} designed an FL-based framework for malicious URL detection in a managed security service provider setting. They looked at various scenarios regarding data partitioning amongst agents and found that the collaborative federated ML model enhanced URL classification performance in all scenarios and improved its rates up to 27\%.

In \cite{thapa2021feddice}, the authors presented the FedDICE framework that integrates FL to SDN-oriented security architecture to enable collaborative learning, detection, and mitigation of ransomware attacks in collaborative environments. This approach achieves performance similar to centralized learning results. 

The work in \cite{demertzis2021blockchained} developed an intelligent threat defense system employing blockchain FL, which seeks to fully upgrade the way passive intelligent systems operate, aiming at intelligently classifying smart cities networks traffic derived from industrial IoT by deep content inspection methods to identify anomalies that are usually due to advanced persistent threat attacks.
\subsection{Summary}
In this section, we present a comprehensive literature review on applying FL at the application and transport layers. There are many improvements and promising solutions that have an impact on the performance of transport, distributed computing, crowdsourcing, cybersecurity, and QoE. The lessons learned from this section are :
\begin{itemize}
    \item Applying FL at both the transport and the application layers introduces a synergistic approach that enhances how data is transported, processed, and utilized across the network. This integration offers the potential to optimize both the technical aspects of data transmission and the functional aspects of user-facing applications.
    \item Ensuring secure data transmission, managing synchronization, optimizing communication overhead, and addressing scalability are considerations that require ongoing attention and innovation.
\end{itemize}
Table \ref{table:6} provides a review of the FL contributions at the APP and the transport layers, highlighting their benefits and drawbacks.

{\renewcommand{\arraystretch}{1.5}%
\begin{table*}[ht!]
\centering
\begin{tabular}{||p{3cm}|p{3cm}|p{5cm}|p{5cm}||}
 \hline
  \textbf{Ref} & \textbf{Application} & \textbf{Benefits} & \textbf{Drawbacks}\\ [0.5ex] 
 \hline
 \cite{pokhrel2020improving,pokhrel2020compound} & Transport performance &  Enhancing TCP performance while preventing privacy leakage & Must implemented in complex networks to analyze the convergence and throughput fairness \\
 \hline
 \cite{sharma2020blockchain,savazzi2020federated}
 & Cooperative and Distributed Computing &  Securely manage distributed and cooperative systems without requiring large computation & Requiring more efficient use of the limited bandwidth, including quantization, compression, or ad hoc channel encoding \\
 \hline
\cite{li2020crowdsfl,zhao2020local,wang2020federated,zhao2021crowdsensing,zhang2021quality,chen2021federated} & Crowdsourcing and Crowdsensing &  Facilitating crowdsourcing and crowdsensing applications with less overhead, less communication cost and higher security & Need to study the waiting for charging as most crowdsensing tasks have temporal deadlines \\
 \hline
 
 \cite{ickin2019privacy,ickin2020ensemble} & QoE & Enabling a seamless QoE model management without having to use sensitive data & Insufficient performance due to the difficulties in accessing QoE datasets \\
 \hline
 \cite{xu2019verifynet,khramtsova2020federated,thapa2021feddice,demertzis2021blockchained}
 & Cybersecurity &  Effectively detecting attacks while addressing issues of data sharing & Have to apply the FL strategy when the attackers influence the model training \\
 \hline
\end{tabular}
\vspace*{0.1cm}
\caption{Summary of FL-based contributions at Transport/APP layers.}
\label{table:6}
\end{table*}}

\section{Federated Learning Applications \& Verticals}
In this section, we discuss the benefits of adopting an FL scheme in wireless network scenarios, including autonomous driving, anomaly detection, industrial operations, smart healthcare, and other applications.
\subsection{Autonomous Driving}
Autonomous driving and autonomous vehicles are now among the most prominent research \cite{hengstler2016applied}. They involve the application of ML and DL techniques that actively select data, transform information, control processes, and make decisions \cite{sallab2017deep}. However, the sequential process of centralized architectures is extremely time-consuming, and the knowledge from the fine-tuned model stays local and not leveraged. Thus, FL is considered to be a strong mechanism that enables devices to interact with their environment and acquire knowledge safely from other devices \cite{elbir2020vehicles}.

To design the autonomous controller of connected and autonomous vehicles (CAV) under the conditions of wireless link uncertainty and environmental dynamics. The work in \cite{zeng2021federated} presented a new DL framework, a dynamic federated proximal algorithm for the training process that introduces a regularizer at the CAVs to minimize the impact of non-IID and unbalanced data on the FL convergence.

In \cite{li2021privacy}, the authors integrated FL into autonomous driving to preserve vehicular privacy and improve model accuracy by keeping original data in a local vehicle and only sharing the training model parameter through the MEC server. The results show that the proposed FL-based autonomous driving system reduces 73.7 \% of training loss, and improves the accuracy to around 5.55 \%.

For privacy-aware and efficient vehicular communication networking, the authors in \cite{pokhrel2020federated} built an autonomous blockchain-based FL system to ensure end-to-end trustworthiness and delay. The proposed framework enables effective autonomous vehicle communication by allowing local on-vehicle ML modules to share and validate their updates in a fully decentralized manner.

The work in \cite{liang2019federated} designed an online federated reinforcement transfer learning process to extract knowledge for autonomous vehicles in real time. In this model, all the participant agents make corresponding actions with the knowledge learned by others, even when they are acting in very different environments. The results show that the proposed approach effectively transfers knowledge online, with better training speed and performance.

To minimize the network-wide power consumption of vehicular users while ensuring high reliability in terms of probabilistic queuing delays, the work presented in \cite{samarakoon2018federated} proposed a novel joint transmit power and resource allocation using FL, which reduces unnecessary overheads.

The authors of \cite{posner2021federated} proposed a new type of FL vehicular network concept namely FVN, which offers more consistent performance than a traditional vehicle network and handles data/computation-intensive applications. The obtained results demonstrate that the proposed framework provides high accuracy and less energy consumption.

\subsection{Anomaly Detection}
FL potentially admits meaningful applications for malicious attack detection in mobile communication systems. For instance, DIoT is an autonomous self-learning distributed system for detecting compromised IoT devices. The approach designed in \cite{nguyen2019diot}, FL efficiently aggregates behavior profiles while relying on device-type-specific communication profiles without human intervention or labeled data. 

In \cite{zhao2019multi}, the authors adopted FL for network anomaly detection and analyzing network traffic to tackle the data scarcity problem and preserve data privacy. It's about designing a multi-task DNN to perform network anomaly detection task, traffic recognition task, and traffic classification task.

An anomaly detection architecture for industrial IoT-based smart manufacturing was presented in \cite{huong2021detecting}. This architecture detects anomalies for time series data typically running inside an industrial system by only sending the training model of each edge to the cloud for the global update. This approach saves 35\% of the bandwidth consumed in the transmission link between the edge and the cloud.

In \cite{zhao2020intelligent}, the authors developed an intelligent intrusion detection framework using FL. First, the LSTM global model is deployed on the servers. Second, each user trains its local model and then uploads its parameters to the central server. Finally, the central server performs the model parameters aggregation to distribute the new knowledge to all participants. This approach achieves higher accuracy and better consistency than conventional methods.

\subsection{Industrial Operations}
Industrial operations refer to the development of many domains, including image processing, robotics, manufacturing, agriculture, and other fields. ML and DL techniques were applied to solve various industrial problems \cite{zellinger2021beyond}. However, traditional centralized learning does not meet the requirements of industrial scenarios because of privacy and scalability concerns. Several works were designed and implemented using FL to address these challenges while generating potential results \cite{savazzi2021opportunities}.

The authors in \cite{hao2019efficient} developed an efficient and privacy-enhanced FL (PEFL) framework to solve various industrial challenging problems in industrial AI. PEFL scheme is non-interactive which prevents privacy leakage from the local gradients and the shared parameters even when an adversary colludes with multiple entities. Results demonstrate the advantage of this approach in terms of accuracy and efficiency.

A verifiable FL (VFL) mechanism with privacy-preserving for big data in industrial IoT was proposed in \cite{fu2020vfl}. This framework effectively ensures the model's security and private gradient by allowing each participant to verify the aggregated results based on Lagrange interpolation, and the verification overhead remains constant regardless of the number of participants. The VFL is applied to many industrial scenarios because of its advantages in terms of verification and total overhead.

To provide a structured collection of requirements and workflows covered in an industrial FL (IFL) architecture, the authors of \cite{hiessl2020industrial} designed an IFL system supporting knowledge exchange in continuously evaluated and updated FL cohorts of learning tasks with sufficient data similarity. The proposed scheme enables optimal collaboration of business partners in common ML problems, prevents negative knowledge transfer, and ensures resource optimization of involved edge devices.

The authors of \cite{sun2020adaptive} proposed a new architecture of digital twin (DT) empowered industrial IoT in which DTs sensitively capture the dynamic changes of the network. The proposed scheme adjusts the aggregation frequency according to the channel state and provides better performance in terms of learning accuracy, convergence rate, and energy saving.

FL offers efficient solutions to collaborative learning in decentralized multi-robot and distributed autonomous systems \cite{xianjia2021federated}. The authors of \cite{zhou2018real} proposed a real-time data processing architecture for multi-robots based on differential FL where a global model is trained on the cloud iteratively and distributed to multiple edge robots. This architecture is applied to multiple robotic recognition tasks and balances the trade-off between performance and privacy.
\subsection{Ultra Reliable Low Latency Communications}
Ultra-reliable low latency communications (URLLC) is one of the key pillars of 5G NR \cite{park2020extreme} for extremely low latency (e.g., 1ms) and high reliability (e.g., 99.999\%). With recent advances in data-driven ML and DL, it is possible to learn a wide range of policies for wireless networks. However, those methods need:
\begin{itemize}
    \item Long training phase.
    \item Large number of training samples.
    \item Long time to obtain enough training samples.
    \item Data access permission.
\end{itemize}
FL is a promising procedure that safely and collaboratively trains ML models and improves URLLC requirements, especially federated reinforcement learning which is one of the distributed setups that learns the environment across numerous decentralized devices without sharing their private data.

A deep reinforcement model was employed in \cite{she2020deep} to develop a multi-level architecture that enables device intelligence, edge intelligence, and cloud intelligence at the user level, cell level, and network level, respectively for URLLC. Considering the computing capacity of each user and each mobile edge computing server is limited, the authors applied FL to improve the learning efficiency.

The work in \cite{samarakoon2019distributed} presented a novel distributed, FL-based, joint transmit power and resource allocation framework to improve URLLC in vehicular communication. The extreme value theory was adopted to characterize the constraints in terms of URLLC by estimating the tail distribution locally without sharing the actual queue length samples. This approach reduces the amount of exchanged data and unnecessary overheads. 

Addressing URLLC's scalability and compatibility with other URLLC/non-URLLC services is a big problem. The authors in \cite{azari2019risk} proposed a hybrid multiple access solution for addressing such issues in a spectrum-/energy-efficient way, by leveraging a distributed hierarchical ML approach for proactive radio resource management to different URLLC traffic streams. Results show the potential of the FL-based scheme that increases 75\% data rate and 99.99\% reliability for both scheduled and non-scheduled traffic types.
\subsection{Tactile Internet}
The tactile internet (TI) is the next step of the internet of things, comprising human-to-machine and machine-to-machine communication. It mainly focuses on providing real-time interactive techniques with a portfolio of engineering, social, and commercial use cases that require a high degree of reliability and latency \cite{simsek20165g}. AI techniques will play an important role in addressing TI challenges \cite{promwongsa2020comprehensive}, especially FL which has a huge potential to optimize energy, resources, and transmission delay for tactile internet applications \cite{mukherjee2020leveraging}.

The authors of \cite{ali2021urllc} used FRL to meet the requirements of B5G mobile communication, especially at the MAC and PHY layers channel access mechanisms that enable URLLC in 5G NR for tactile internet. The FRL is an emerging solution to handle emerging wireless network scenarios since it learns from unlabeled data and avoids the expectation of a formulated Markov decision process to make optimal decisions, routing, and link selection.
\subsection{Virtual/Augmented/Extended Reality \& Metaverse}
Virtual reality (VR) is a computer-generated environment with realistic-looking scenes and objects that give the user the feeling of being completely immersed in their surroundings. Deploying this new technology over wireless networks is an essential stepping stone towards flexible deployment of pervasive VR applications \cite{bastug2017toward}. Minimizing the occurrence of the break-in presence (BIP) is one of the key challenges in VR that detaches the users from their virtual world. The authors of \cite{chen2019VR} designed a novel model that jointly considers the VR application type, transmission delay, VR video quality, and users’ awareness of the virtual environment to measure the BIP for wireless VR users. They adopted an FL strategy to enable multiple BSs to locally train their model using the gathered data and cooperatively build a global model to predict the entire users’ locations and orientations. Compared to centralized approaches, this method achieves significant performance by reducing the users’ BIP.

Augmented reality (AR) is an enhanced version of the real physical world that is achieved using digital visual elements, sound, or other sensory stimuli delivered via technology. Driven by computer vision and AI tools, AR technology has shown a strong momentum of development \cite{chen2019overview}. However, the bandwidth available to transmit and process large amounts of generated data is extremely limited, making it extremely difficult for AR to detect and classify objects and achieve their goals of perfectly combining the corresponding virtual contents with the real world. To solve the classification challenge for latency-sensitive AR applications that require a high data rate, the work in \cite{chen2020federated} proposed a framework that combines FL with mobile edge computing (MEC) to obtain the global optimal ML model. Evaluations show that the presented approach requires significantly fewer training iterations compared to traditional centralized frameworks. To solve the privacy issues in AR, the authors of \cite{wangoutput} proposed a multi-user AR output strategy model using a hierarchical federated reinforcement learning method for generating and aggregating the AR output strategy model for multiple users. This framework effectively learns the information of multi-user AR scenarios and improves the robustness of the adaptive output strategy model.

Extended reality (XR) refers to all real and virtual environments and human-machine interactions generated by computer technology and wearable devices \cite{ratcliffe2021extended}. XR technologies collect and process huge amounts of detailed and personal data about users, FL is a promising method to handle such challenges while reducing costs and preserving dispersed datasets. The authors in \cite{barbieri2021decentralized} investigated distributed FL methods in V2X networks for augmenting the capability of road user/object classification based on Lidar data, where the model sharing is implemented on a large number of layers, depending on the required efficiency and bandwidth requirements.

The metaverse refers to a virtual shared space, often encompassing AR, VR, XR, and other immersive digital environments. It's an interconnected digital universe where users interact, socialize, work, and engage in a wide range of activities \cite{chehimi2023roadmap}. Applying FL to the metaverse protects the data privacy of clients and reduces the need for high computing power and
high memory on servers \cite{chen2023federated}.
\subsection{Smart Healthcare}
ML and DL techniques have shown potential for accelerating the advancement of smart healthcare. However, in several healthcare situations, the limited amount of data available does not allow for building very powerful models to investigate the existing challenges. To this end, FL is employed to allow medical institutions to share their experiences, and not their data, with a guarantee of privacy to safely perform personalized healthcare and biomedical tasks \cite{nguyen2022federated,xu2021federated}.

The authors of \cite{chen2020fedhealth} developed the FedHealth framework using FTL that aggregates the data from the distributed organizations to train powerful ML algorithms while preserving client privacy. Following the initial training of a global model in the cloud, FedHealth leverages transfer learning to create personalized models for each organization. These personalized models are continually updated by transmitting the modified global model parameters back to the respective organizations.

In \cite{lim2020dynamic}, the authors proposed an edge computing-empowered FL-based system model to facilitate privacy-preserving collaborative model training across distributed participants for the development of healthcare applications. The result of the present approach shows that the contract design is self-revealing across two time periods and returns higher profits for the model owner as compared to the uniform strategy.

The clustered FL was adopted in \cite{qayyum2021collaborative} to develop an automatic diagnosis system for the COVID-19 pandemic. This approach allows remote healthcare institutions to benefit from collaborative learning without sharing local data and associated information.
\subsection{Recommendation Systems}
Recommendation systems in wireless networks play a significant role in enhancing user experiences, optimizing network resource allocation, and improving overall network efficiency \cite{fu2023new}. Implementing efficient recommendation systems in wireless networks requires a combination of data analytics, ML algorithms, and real-time data processing. This encompasses the collection and analysis of user behavior, network performance metrics, and contextual data \cite{gashi2023secure}. 
FL has recently been applied to recommendation systems to protect user privacy and boost prediction accuracy \cite{vaishnavi2023three}. In FL settings, recommendation systems train recommendation models by collecting the intermediate parameters instead of the real user data, which greatly enhances user privacy \cite{yang2020federated}. A privacy-preserved recommender system framework was proposed in \cite{qin2021novel} by integrating FL architecture to enable the recommendation algorithm to be trained and carry out inference without centrally collecting users’ private data. This approach reduces the privacy leakage risk, satisfies legal and regulatory requirements, and allows various recommendation algorithms to be applied.
{\renewcommand{\arraystretch}{1.5}%
\begin{table*}[ht!]
\centering
\begin{tabular}{||p{3cm}|p{3cm}|p{5cm}|p{5cm}||}
 \hline
  \textbf{Ref} & \textbf{Application} & \textbf{Benefits} & \textbf{Drawbacks} \\ [0.5ex] 
 \hline
 \cite{zeng2021federated,li2021privacy,pokhrel2020federated,liang2019federated,samarakoon2018federated,posner2021federated} & Autonomous driving &  
 Achieving potential learning across several decentralized vehicles by only sharing local model parameters & Need to limit the increase in the overall delay and avoid the privacy leakage risk \\
 \hline
 \cite{nguyen2019diot,zhao2019multi,huong2021detecting,zhao2020intelligent}
 & Anomaly detection &  Improving the scalability of malicious attacks detection for distributed wireless communication networks & The training phase should be optimized in terms of weights communicating and performance \\
 \hline
 \cite{she2020deep,samarakoon2019distributed,azari2019risk}
 & URLLC & Decreasing service delay and providing high reliability while preserving user's information & Cannot handle the non-IID training data and a large number of connected devices\\
 \hline
 \cite{ali2021urllc} & Tactile internet & Achieving intelligent and optimized network control and resource allocation for URLLC requirements & Each device must learn the optimal decision with the help of exploration\\
 \hline
 \cite{chen2019VR,chen2020federated,wangoutput,barbieri2021decentralized} & VR/AR/XR and Metaverse &  Improving the generation of virtual objects by optimizing the resource management and reducing the overall latency for VR/AR users & The security problems caused by non-visual output along with the conflict between multiple output types\\
 \hline
 \cite{chen2020fedhealth,lim2020dynamic,qayyum2021collaborative} & Smart healthcare
 & Allowing organizations to benefit securely from distributed medical experiences & Must explore the performance in varied distributed data to enhance efficiency and flexibility \\
 \hline
\cite{gashi2023secure,vaishnavi2023three,yang2020federated,qin2021novel} & Recommendation system
 & Training a distributed recommender system while protecting user privacy & The communication between the user and the server leads to increased overhead and latency \\
 \hline
\end{tabular}
\vspace*{0.1cm}
\caption{Summary of FL-based verticals and applications.}
\label{table:7}
\end{table*}}
\subsection{Summary}
In this section, we have presented a comprehensive summary of the latest research related to FL in 6G systems. We have focused on highlighting the pivotal enhancements gained through the adoption of FL in contrast to centralized ML. A summarized overview is found in Table \ref{table:7}.

\section{Insights and Open Problems}
FL has gained tremendous interest in wireless communication networks, providing better and safer methods that effectively enhance mobile operations. Lately, numerous substantial FL-based advances in wireless systems have been developed promising a wide deployment in future 6G. However, FL is still a nascent research with several challenging issues such as to privacy and security risks, data scarcity/quality, system heterogeneity, complexity, and convergence \cite{huang2022learn}. In this section, we discuss open problems of FL and provide a summary of recent solutions, insights, and opportunities.
\begin{figure}[t]
\centering
\includegraphics[width=\columnwidth]{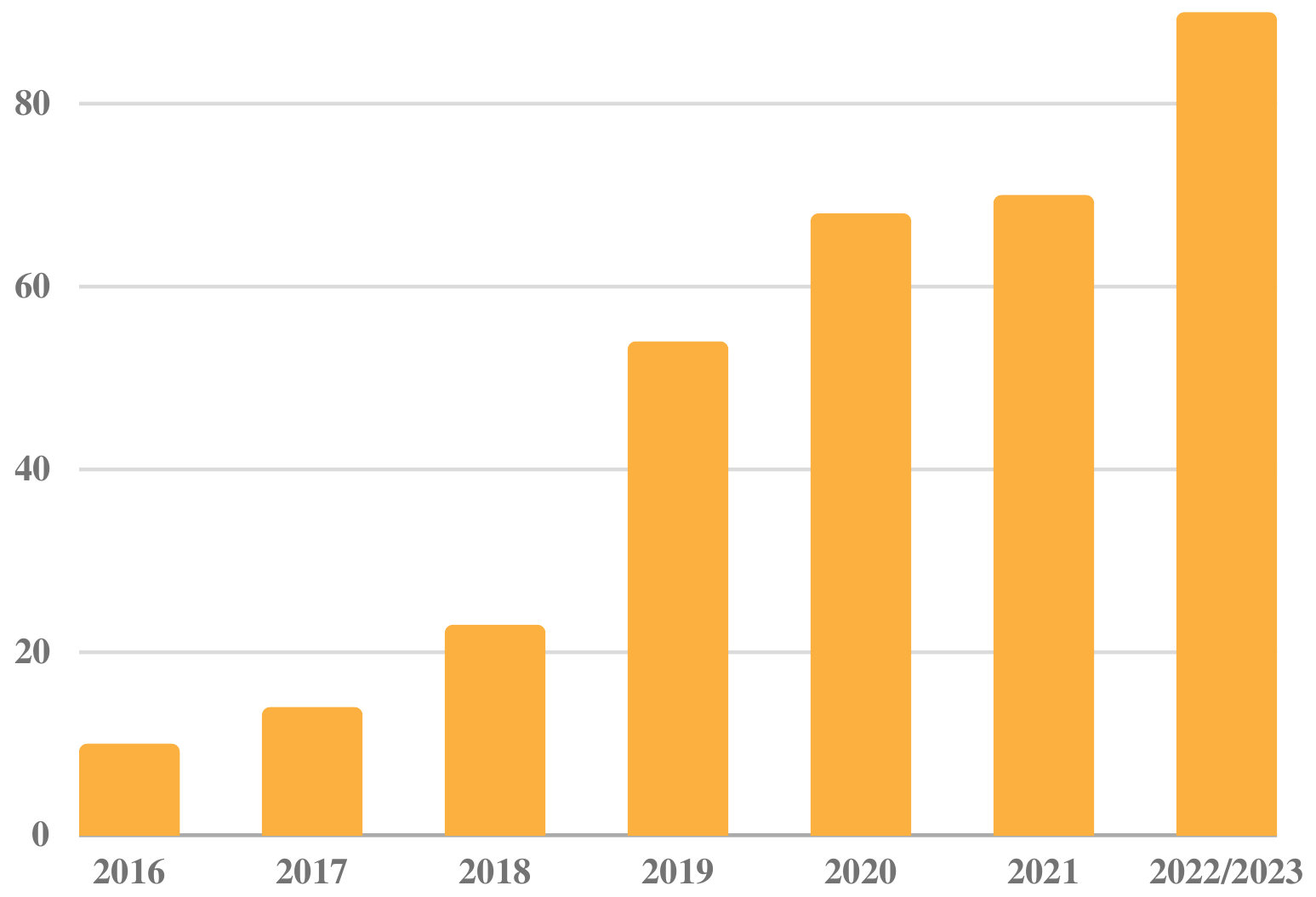}
\caption{Year-wise publications of FL for mobile networks [generated from all cited papers].}
\label{bar chart}
\end{figure}
\begin{figure}[ht]
\centering
\includegraphics[width=0.935\columnwidth]{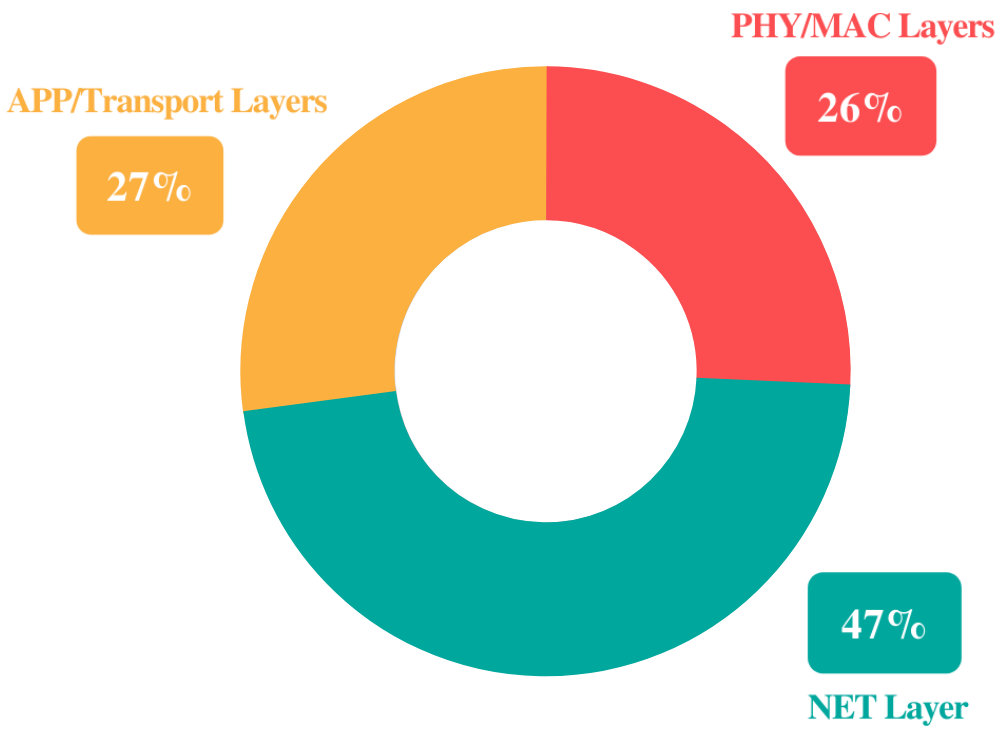}
\caption{Proportion of FL-based approaches per network layer [generated from all cited papers].}
\label{pie chart}
\end{figure}

\subsection{Lessons Learned}
In this tutorial, we compiled, classified, and carefully analyzed recent results and pioneering research breakthroughs related to FL. As illustrated in Fig. \ref{bar chart}, the number of FL learning contributions in mobile networks has grown steadily over the past seven years, and widely applied to network design, as visualized in Fig. \ref{pie chart}. Indeed, FL leverages efficient and effective solutions compared to traditional AI techniques to handle wireless problems while preserving privacy. To sum up, the lessons learned from our study are:
\begin{itemize}
    \item 6G and beyond call for high network flexibility, reliability, privacy, and intelligence. Those desired features are efficiently and effectively provided by FL approaches.
    \item FL has the potential to shape next-generation wireless systems by enhancing the protocol stack and meeting future demands of data-intensive applications. By incorporating FL across multiple network layers, wireless systems will benefit from fundamental values such as sustainability, trustworthiness, fairness, reduced communication overhead, high resilience and neverseen degree of versatility. 
    \item FL support by wireless networks has the potential to revolutionize several industries, from healthcare and manufacturing to smart cities and smart farming, as it enables novel personalized services, efficient resource management, and data-driven decision-making.
    \item FL is speculated to be a transformative approach for the evolution of future wireless communication systems. However, challenges like communication efficiency, model aggregation, and security still need to be addressed to fully harness FL potential in 6G and future wireless networks. Integrating FL into the design loop of future networks will redefine their capabilities in terms of performance, privacy, resilience, and intelligence.
\end{itemize}

\subsection{Federated Learning Challenges}
Through our systematic review of existing literature, we have identified some primary bottlenecks that dominate current FL research:

\subsubsection{Privacy and Security}
Contrary to decentralized schemes, federated learning was designed to provide a secure machine learning among distributed learning agents. However, data privacy is still challenging due to the presence of attackers such as on-device malicious clients to malicious users with only a black-box access to the model \cite{lyu2020threats}. Although local data might not escape the device, an intruder still can learn the presence of a data point used for training \cite{wen2023survey}. Many efforts have been spent to deal with this issue using transfer ML, secure aggregation, and attack recognition \cite{fereidooni2021safelearn,zhang2021secure}.

\subsubsection{Complexity and Convergence}
Federated learning suffers from high computational complexity during training due to running consensus algorithms over heterogeneous massive environments over perturbed/faulty communication systems. It includes a large number of handsets with data generated ruled by diverse distributions, which results in high communication overhead. The goal of FL is to train high-quality models with lower communication complexity and faster convergence, hence it is crucial to design efficient schemes to overcome such a limitation. In \cite{konevcny2016federated1}, two solutions were designed to reduce uplink communication costs: learning an update directly from a reduced set of variables, and learning a complete model update and compressing it before sending it to the server. Based on gradient descent, the authors of \cite{wang2019adaptive} analyzed the convergence bound for FL with non-IID datasets. Other researches were proposed for uplink and downlink transmission to improve the convergence time \cite{haddadpour2019convergence,dinh2020federated,wei2021federated,chen2021communication}.

\subsubsection{Client Selection}
FL clients execute limited number of training iterations using local update parameters and only exchange model updates with the aggregating server. The number of agents is sufficiently large and the bandwidth available for model update distribution is limited. Thus, it is practical to select the best subset of clients in the training stage, as client selection policy is critical to FL in terms of training efficiency and global model quality. Manifestly, the client nodes generate highly heterogeneous local datasets and handle different computation capabilities \cite{ahmadi2022dqre}, which affects the global model convergence. The FL framework has to be robust to devise dropout and anticipate the scenarios where only a small number of participants are left connected to participate in a training iteration. Several research activities were devoted to bridge this gap. In \cite{cho2020client}, the authors design a client selection framework that flexibly spans the trade-off between convergence speed and solution bias. They demonstrate that biasing client selection towards clients with higher local loss achieves faster error convergence. Tang et al. \cite{tang2021fedgp} use the correlations between clients to achieve a faster and more stable convergence in heterogeneous scenarios. Furthermore, in \cite{nishio2019client}, the authors propose a client selection mechanism that allows the server to pick the clients that can participate based on their available resources. The server decides on the amount of data, energy, and CPU resources required to take part in each round. In \cite{zhang2023delay}, a delay-constrained client selection framework is proposed for heterogeneous FL in intelligent transportation systems, aiming to improve the model accuracy, training speed, and transmission time. Further, reinforcement learning is deployed to improve client selection performance by involving a reinforcement learning agent that learns a client selection policy \cite{cheng2023learning}. Also, a multi-armed bandit model is used in \cite{shi2023efficient} for client selection to boost volatile FL by speeding up model convergence, promoting model accuracy, and reducing energy consumption. Alternatively, \cite{kang2023ga} and \cite{chahoud2023demand} propose a client selection method using a genetic algorithm based on the client’s cost and the model accuracy. In \cite{tang2023stackelberg}, the authors present the client selection as a game where the clients express their interest in participating in the FL process.

\subsubsection{Fairness \& Role Distribution}
FL systems are designed to train distributed models using decentralized and private local training data, which requires to assess the mid/long-term training robustness and resilience. Most of current techniques combine models and prioritize them proportionately to the frequency of local samples. However, the question of role distribution between FL participants remains unsolved and calls for implementing high fairness, this was examined from the perspective of ensuring accuracy across clients \cite{michieli2021all}. Numerous works employ robust approaches to address fairness, including \cite{hao2021towards,huang2020efficiency,lyu2020collaborative}, the authors tried to improve the overall fairness without compromising the predictive performance using zero-shot data augmentation, Lyapunov optimization, and participants reputation, respectively. The research preented in \cite{wang2021fair} introduces federated fair averaging to mitigate potential conflicts among clients before averaging their gradients. In \cite{chahoud2023demand}, the authors propose an FL framework that uses a genetic algorithm to ensure fairness and improve the volume and heterogeneity of data in the FL process. Another approach for fair client selection is presented in \cite{he2023three}, which uses a dynamic multi-criteria scheme to promote fairness among clients in each round of the FL process.

\subsubsection{Data quality}
The data quality in FL is a critical factor that directly impacts the accuracy, reliability, and performance of the trained models. Maintaining high data quality for FL is a multifaceted challenge that requires careful considerations:
\begin{itemize}
    \item \emph{Data heterogeneity:} In real-world, i.i.d. assumption does not hold as data sources may experience spatial or temporal correlations, which may lead to convergence or accuracy inefficiencies and serious fairness issues. Data heterogeneity is mainly due to variations in data distribution, format, and quality. Addressing these differences and ensuring representative data from all devices is crucial reliable training.
    \item \emph{Data labeling:} For supervised learning tasks, inconsistent labeling across devices may lead to biased models and erroneous predictions.
    \item \emph{Imbalanced data:} Handling data imbalance in FL is essential to ensure generalization and fair representation across all classes.
    \item \emph{Data freshness:} Ensuring data freshness is crucial to capture recent trends, patterns, real-time changes, and user behaviors. Meanwhile, network latency, communication overhead, and battery consumption are to be addressed for efficient data freshness and communication trade-offs.
    \item \emph{Data Regulation:} To comply with data storage/processing location, protection regulations, and other constraints, FL systems have to consider legal and ethical aspects in the whole data value chain.
\end{itemize}
The framework introduced in \cite{peyvandi2022privacy} combines FL and blockchain to ensure high data quality for complex ML tasks while increasing the accuracy of the trained models. Moreover, employing pre-processing techniques, local validation, and rigorous aggregation strategies contribute to maintaining data quality and fostering more robust and dependable FL models.

\subsection{Wireless Challenges for Federated Learning}
To train FL algorithms, all participants must transmit their training parameters over a shared network which require to face several important communication issues \cite{singh2022federated}. In this section, we present the main problems related to the implementation of FL over wireless environments.

\subsubsection{Resource Constraints}
Wireless devices are usually equipped with limited hardware and software capabilities, which may drastically affect the local model training efficiency and the whole FL feasibility. Namely, the FL may experience a loss of efficiency due to the following specific aspects: 
\begin{itemize}
    \item \textbf{Computational power:} Most of mobile handsets and IoT devices have limited computational capabilities, which affects their ability to perform complex model updates and heavy local training. This way, the so called stragglers with low processing capabilities may lower the training quality and slow the overall federated learning process. Therefore, it is fo paramount importance to elaborate policies to handle stragglers, such as strategic redundancy or per-client weighting.
    \item \textbf{Memory limitations:} The low amount of memory installed on devices restricts the size of the models they can handle and increase the processing time. Therefore, only lightweight model training and/or low-moderate accuracy may be supported.
    \item \textbf{Wireless context:} Transmitting model updates and gradients over wireless requires efficient bandwidth allocation to deal with the channel variability. Devices suffering from uncontrolled/unpredictable wireless context may experience poor communication experience, leading to unbounded delays and communication overhead in particular under large models or frequent updates.
    \item \textbf{Energy Considerations:} Running mobile handsets and IoT sensors faces a serious energy constraints due to limited battery life. Indeed, intensive model training tasks may lead to rapid battery depletion, affecting the FL robustness and user experience afterwards. Then, it is of utmost importance to design energy-efficient model update mechanisms and green communication strategies to sustain a satisfactory learning experience. 
\end{itemize}
Abovementioned limitations impact the effectiveness, efficiency, and feasibility of implementing FL over wireless network systems. Numerous works (e.g., \cite{konevcny2015federated,chen2020joint}) explored techniques such as model compression, federated optimization strategies, and adaptive learning algorithms tailored to resource-constrained devices to balance the model accuracy and resource efficiency while enabling effective participation of a diverse range of devices in the FL ecosystem.

\subsubsection{Federated Learning Over Wireless} 
Wireless links introduce additional challenges when transmitting model updates and data from/to participating devices to/from the central server in an FL framework. Symbol errors introduced by the unreliable nature of the wireless channel impact the quality and correctness of the FL updates among users and affect the performance of FL algorithms, as well as their convergence speed. To address the inherently unreliable connectivity in FL, the authors of \cite{yang2019scheduling,chen2020convergence} consider error correction coding, adaptive modulation and coding schemes, robust aggregation algorithms, and the use of redundant data transmission. These approaches aim to ensure that the FL process continues running effectively even under faulty wireless communications.

\subsubsection{Over-the-Air Data Aggregation} 
Over-the-air computation offers the advantage of low resource consumption since the BS only needs to handle functions uploaded by users rather than dealing with individual data. FL implementing over-the-air aggregation enables users to share the same spectral resources, resulting in communication efficiency enhancement \cite{xiao2023over}. However, several works such as \cite{oksuz2023federated,tegin2023federated} have examined the impact of the channel time variations on the convergence of the FL with over-the-air aggregation, and show that the resulting undesired interference has only limited destructive effects, which do not prevent the convergence of the learning algorithm. Furthermore, other issues should be addressed such as model distortion caused by channel fading, inefficient over-the-air aggregation of locally trained models with significant data imbalances, and the constrained availability and verification of individual local models.

\subsection{Hot Topics and Insights}
In this section, we provide a broad overview of the hot topics that still need addressed for meaningful FL. We also outline several directions of future work with continuingly high interest for a wide range of research communities.
\subsubsection{FL and Edge Computing/Networking} Combining FL and edge computing promises to handle the massive amount of data securely gathered from ubiquitous mobile devices. Edge computing refers to situations in which user nodes have computational and/or storage capability but need edge help for coordination or to offload the most computationally intensive activities. FL has long been recognized as an excellent match for edge-hosted applications, and numerous research efforts have aimed to make it as efficient as possible \cite{malandrino2021federated}. Nevertheless, the interaction between FL and edge computing/networking is becoming an open battlefield with several issues to solve. First, the nodes connect to the edge-based server using different radio access technologies, which influences the delay experienced while delivering model updates. Second, heterogeneous mobile clients with different capability also affects the FL performance. Therefore, the set of participating clients in the FL process needs to be strategically chosen to communicate effectively. Third, during the exchange of model updates, several clients participate by contributing their on-device training data, and it is difficult to prevent malicious clients from sending fake data to poison the training process which in turn poisons the model or the received model gradient parameters.
\subsubsection{FL for 6G applications}
Future 6G applications are expected to generate and handle large amounts of data, such as extended reality (XR), mixed reality (MR), wireless brain-computer interactions, hologram communications, tactile internet, mURLLC, XURLLC, and other use cases that require high QoE, astonishing reliability and extremely low latency \cite{tan2022towards}. Federated ML schemes are highly recommended to obtain greater learning and secure communication between distributed devices \cite{barbieri2022decentralized}. However, new concerns arise when applying FL for emerging technologies, essentially the convergence analysis that is affected by the large propagation attenuation in THz and the optimization of the parameters for quantum key distribution to name a few.
\subsubsection{FL meets Blockchain}
Blockchain technology combines encryption with distributed computing to improve the privacy of decentralized networks while cutting down the need for a central authority. In \cite{singh2023fusionfedblock,qu2022blockchain}, it has beeb shown that implementing FL and blockchain helps automating energy transfers and allows autonomous energy providers to safely trade value \cite{kim2019blockchained}. Additionally, game theoretic tools may help FL to substantially reduce communications among distributed wireless networks and address the fundamental problems of large-scale distributed and privacy-preserving learning algorithms \cite{ma2020federated, mehrjou2021federated}. Local federated learners are the players and the aggregation of the received gradients in a central server is the mean-field game effect. Adopting incentive mechanisms via blockchain and game theory helps distributed systems exploring the user's habits and preferences, assessing and improving the relevance of online search engines.
\subsubsection{Federated Distillation}
Despite being communication-efficient, FL still requires the transmission of extensive models from large number of clients through the air. FL
has well-known limits, such as:
\begin{itemize}
    \item Clients must implement the same model architecture;
    \item Transmitting the huge model weights and updates implies high communication costs;
    \item Parameter-averaging aggregation schemes perform poorly due to client model drifts. 
\end{itemize}
Knowledge distillation helps overcoming limitations of the parameter-averaging technique \cite{jeong2018communication}. Federated distillation is a process of distilling knowledge from a larger central model to a smaller model on edge devices \cite{mora2022knowledge}. Within this process, knowledge is transferred between models by having each model acting as a teacher to the other models. Each model shares its knowledge with the other models, allowing the models to learn from each other and improve their performance \cite{chen2021distributed}. Such a mechanism leads to improved accuracy as more data is implicitly used for training. It also offers improved privacy and security as data is not shared. Moreover, federated distillation exhibits  a reduced bandwidth usage due to the distributed training. Likewise, the authors of \cite{le2021distilling} propose to distill knowledge in the FL process, with key idea relying on exchanging soft targets instead of transferring the model parameters between server and clients. The authors showed that their solution can reduce both communication overhead and computation costs.
\subsubsection{FL for Semantic Communications}
Semantic Communication (SC) is a new paradigm, expected to be a major part of 6G and beyond. It allows to achieve more efficient data processing and transmission considering user expectation and application requirements as well as the semantic/meaning of the data being sent. SC combines natural language processing, AI, and ML to create a powerful and dynamic communication system able to understand the context of the data being processed. Indeed, it allows to build more secure, reliable, and energy-efficient communication networks. In addition, semantic communication is expected to reduce the latency and complexity of data processing, allowing for more efficient data transmission and improved user experience \cite{yang2022semantic}. This way, SC is anticipated to support novel applications and services, such as intelligent virtual assistants, smart robots, and autonomous vehicles. However, some SC processes such as semantic detection, knowledge modelling, and coordination, are resource-hungry and may show inefficient behaviors, especially for communication between a source and a destination.

FL framework addresses this challenge by training a shared model to process the common semantic knowledge without exposing local semantic data from each edge server. Instead, the edge servers collaborate by exchanging intermediate model training results. In \cite{shi2021semantic}, a federated ML is used to support resource-efficient semantic-aware networking. The proposed solution allows each user to offload computationally intensive semantic encoding and decoding tasks to edge servers, while it protects its own model-related information by coordinating via intermediate results.
\subsubsection{FL and Energy Efficiency}
Energy efficiency is critical to build sustainable 6G and next generations of mobile networks. Inline with this goal, the energy consumption of FL during the training process \cite{shi2022toward} must be carefully optimized and underlying schemes must be energy-aware designed. It is understood, FL is more energy-efficient than traditional centralized learning approaches, but there are still several energy wasting operations to be rethought for green learning \cite{yang2020energy}. First, the distributed nature of FL requires more computing resources, which may results in increased energy consumption. Second, FL might require data to be transferred across devices/edge/cloud, leading to increased energy consumption. Finally, the FL process typically requires multiple rounds of data transmission and computation, which further increases the overall energy consumption. In \cite{khowaja2021toward} a distributed FL framework is designed to overcome the connectivity and energy depletion for distant devices. The framework is suitable for mobile edge computing that connects the devices in a distributed manner using clustering protocols. This approach resolves centralized systems issues and exhibits scalable, faster, and energy-efficient communication when exchanging the trained models.

In the near future, FL is trusted to run large neural networks or similar algorithms at mobile devices and client handsets level. This is not feasible for resource-constrained devices. In this regard, one major challenge is how to design FL algorithms to minimize the use of computing and communication resources. One potential perspective is to efficiently use quantization \cite{kim2023green,kim2022tradeoff}, which was shown to significantly reduce energy consumption. Indeed, there is a need for new frameworks to build green FL algorithms that achieve both high computational and communication efficiency, while requiring low resource and energy budget.

\section{Conclusion}
Federated learning shows great potential in addressing privacy concerns and communication costs in various sectors, especially in current and upcoming mobile network generations. This paper thoroughly examines FL's role in wireless communication networks, comparing different ML architectures and exploring the FL fundamentals, techniques, and frameworks. We also review the latest contributions that use FL to improve communication and KPIs within the protocol stack. Finally, we discuss important insights and challenges that come with deploying FL strategies in 5G, 6G, and beyond. Overall, FL is a vital research domain with far-reaching implications for the future, and this paper provides a detailed analysis of its potential and  impact.

\section*{Table of Abbreviations \& Acronyms}
\noindent The main abbreviations used in this survey are listed bellow:\\

\small
\begin{tabular}{l l}
 \textbf{Abbreviation} & \textbf{Description} \vspace{0.3cm}\\
 5G & Fifth Generation Mobile Network \\
 6G & Sixth Generation Mobile Network \\
 NR & (5G) New Radio \\
 RF & Radio Frequency \\
 AI & Artificial Intelligence\\
 FL &  Federated Learning \\ 
 ML & Machine Learning \\
 DL & Deep Learning \\
 FRL & Federated Reinforcement Learning \\
 OSI & Open Systems Interconnection \\
 PHY & Physical Layer\\
 NET & Network Layer \\
 MAC & Medium Access Control Layer \\
 APP & Application \\
 IoT & Internet of Things \\
 CNN & Convolutional Neural Network\\
 NN & Neural Network \\
 HFL & Horizontal Federated Learning\\
 VFL & Vertical Federated Learning\\
 FTL & Federated Transfer Learning\\
 MMVFL & Multi-class Vertical Federated Learning\\
 HFCL & Hybrid Federated Centralized Learning \\
 FedAvg & Federated Averaging \\
 SimFL & Similarity Federated Learning \\
 TFF & TensorFlow Federated  \\
 FATE & Federated AI Technology Enabler \\
 FedML & Federated Machine Learning \\
 FedRec & FL-based receiver \\
 AMC & Automatic Modulation Classification \\ 
 TCP & Transmission Control Protocol \\
 CSI & Channel State Information \\
 RF & Radio Frequency \\
 RIS & Reconfigurable Intelligent Surfaces\\
 FED-WG & Federated Weights and Gradients \\
 MIMO & Multiple Input Multiple Outputs \\
 mm-Wave & millimeter Wave \\
 SPIM & Spatial Path Index Modulation \\
 NOMA & Non-orthogonal Multiple Access \\
 MTD & Machine Type Devices \\
 MEC & Multi-access edge computing \\
 LSTM & Long Short Term Memory \\
 SVM & Support Vector Machine \\
 UAV & Unmanned Aerial Vehicle \\
 QoE & Quality of Experience \\
 HDRL & Hybrid FL Reinforcement Learning \\
 VR & Virtual Reality \\
 AR & Augmented Reality \\
 XR & Extended Reality \\
 TI & Tactile Internet \\
 URLLC & Ultra-Reliable Low-Latency Communication \\
 TC & Traffic Classification \\
\end{tabular}
\bibliography{references.bib}
\bibliographystyle{IEEEtran}
\textbf{Maryam Ben Driss}  (Student Member, IEEE) received the B.Sc. degree in Computer Science in 2018 and the M.Sc. degree in Big Data and Data Science in 2020 from the University of Hassan II, Faculty of Sciences Ben M’Sik, Casablanca, Morocco. She is currently pursuing her Ph.D. degree at the National Higher School of Electricity and Mechanics (ENSEM) in Casablanca. Her research interests include Artificial Intelligence, Federated Learning, Machine Learning, Deep Learning, Cellular Networks, 5G, 6G and beyond.\\

\noindent\textbf{Essaid Sabir} (Senior Member, IEEE) received the Ph.D. degree (Hons.) in networking and computer engineering from Avignon University, France, in 2010. He has been a non-tenure-track Assistant Professor at Avignon University, from 2009 to 2012. He has been a Professor at  Hassan II university of Casablanca until late 2022, where he was leading the NEST research Group. He is a professor with the department of computer science, Universit\'e du Qu\'ebec \`a Montr\'eal. His research interests include 5G/6G, wireless networks, IoT, AI/ML, and game theory. His work has been awarded in four international conferences. To bridge the gap between academia and industry, he founded the International Conference on Ubiquitous Networking (UNet) and co-founded the WINCOM conference series. He serves as a guest editor for many journals. He organized numerous events and played executive roles for other major events.\\

\noindent\textbf{Halima Elbiaze} (Senior Member, IEEE) received the B.S. degree in applied mathematics from the University of MV, Morocco, in 1996, the M.Sc. degree in telecommunication systems from the University of Versailles, in 1998, and the Ph.D. degree in computer science from Institut National des Télécommunications, Paris, France, in 2002. Since 2003, she has been with the Department of Computer Science, Université du Québec à Montréal, QC, Canada, where she is currently an Associate Professor. She has authored or co-authored many journal and conference papers. Her research interests include network performance evaluation, traffic engineering, and quality of service management in optical and wireless networks.\\

\noindent\textbf{Walid Saad} (Fellow Member, IEEE) received his Ph.D. degree from the University of Oslo in 2010. He is a Professor at the Department of Electrical and Computer Engineering at Virginia Tech where he leads the Network Science, Wireless, and Security (NEWS) laboratory. His research interests include wireless networks, machine learning, game theory, cybersecurity, unmanned aerial vehicles, semantic communications, and cyber-physical systems. He was the author/co-author of eleven conferences' best paper awards and the 2015 and 2022 IEEE ComSoc Fred W. Ellersick Prize.
\end{document}